\documentclass{article} 
\usepackage{TR-CS-UOM-2023,times}
\usepackage{booktabs}       
\usepackage{amsfonts}       
\usepackage{nicefrac}       
\usepackage{microtype}      
\usepackage{xcolor}         
\usepackage{amsfonts}
\usepackage{caption}
\usepackage{graphicx}
\usepackage{mathtools}
\usepackage{subcaption}
\usepackage{algorithm}
\usepackage{algpseudocode}
\usepackage[colorlinks,urlcolor=blue,linkcolor=blue,citecolor=blue]{hyperref}
\usepackage{soul}
\definecolor{lisheng}{rgb}{0.2, 0, 1}

\definecolor{highlight}{rgb}{0, 0, 1}
\newcommand{\highlight}{\textcolor{highlight}}
\usepackage{amsthm}
\usepackage{enumitem}
\usepackage{textcomp}
\theoremstyle{definition}
\newtheorem{definition}{Definition}

\newcommand{\btheta}{{\bar{\theta}}}
\newcommand{\bpsi}{{\bar{\psi}}}
\usepackage[utf8]{inputenc} 
\usepackage[T1]{fontenc}    
\usepackage{url}            
\usepackage{titlesec}

\title{Bias Resilient Multi-Step Off-Policy\\ Goal-Conditioned Reinforcement Learning}

\author{Lisheng Wu, Ke Chen \\
Department of Computer Science\\
University of Manchester\\
Manchester, M13 9PL, UK \\
\texttt{\{lisheng.wu,ke.chen\}@manchester.ac.uk} \\
}

\iclrfinalcopy 
\begin{document}

\maketitle

\begin{abstract}
In goal-conditioned reinforcement learning (GCRL), sparse rewards present significant challenges, often obstructing efficient learning. Although multi-step GCRL can boost this efficiency, it can also lead to off-policy biases in target values. This paper dives deep into these biases, categorizing them into two distinct categories: ``shooting" and ``shifting". Recognizing that certain behavior policies can hasten policy refinement, we present solutions designed to capitalize on the positive aspects of these biases while minimizing their drawbacks, enabling the use of larger step sizes to speed up GCRL. An empirical study demonstrates that our approach ensures a resilient and robust improvement, even in ten-step learning scenarios, leading to superior learning efficiency and performance that generally surpass the baseline and several state-of-the-art multi-step GCRL benchmarks.
\end{abstract}
\section{Introduction}
In goal-conditioned reinforcement learning (GCRL) \citep{schaul2015universal,   liu2022goal}, agents aim to complete various tasks each characterized by different goals. A primary challenge arises from the complexity of reward engineering, often leading to sparse reward settings—where valuable rewards are only given upon reaching the goals. Such limited feedback inhibits efficient learning, causing delays in the learning progression. Although relabeling goals based on actual outcomes for each trajectory augments the learning experience \citep{andrychowicz2017hindsight}, the slow propagation of these sparse learning signals remains a bottleneck for the rapid acquisition of successful policies. To counteract this, multi-step hindsight experience replay (MHER) \citep{yang2021bias} accelerates the relay of relabeled rewards using multi-step targets.
However, in the off-policy setting, multi-step learning can introduce off-policy biases originating from the divergence between target and behavior
policies.
 As the step-size magnifies, these challenges are further amplified. While methods like MHER($\lambda$) and MMHER, proposed by \cite{yang2021bias}, attempt to mitigate these biases, we observe that they lack the requisite resilience, rendering their methods effective only for smaller step sizes.

Though off-policy bias in multi-step RL has been well-studied \citep{sutton2018reinforcement,munos2016safe,de2018multi}, its nuances within GCRL present novel challenges. In GCRL, the mere act of reaching a goal does not guarantee its sustained achievement. The agent must learn the optimal strategy to consistently stay on the goal, which necessitates continuous bootstrapping of value functions on goal states. As off-policy biases accumulate at these goal states, they can retroactively distort the value estimates of preceding states. In light of this, we categorize off-policy bias into the \textit{shooting} bias that accumulates up to the goal states and introduce the novel concept of \textit{shifting} bias that builds within the goal states.
While off-policy biases are typically perceived as detrimental, by examining the multi-step target values, we propose that certain behavior policies offer superior alternatives over the multi-step reward accumulation, guiding the agent to refine its policy more swiftly. Building on this understanding, we present tailored solutions that harness the positive effects of both biases while curbing their adverse effects.

In the broader context of examining bias problems in multi-step GCRL, it is imperative to recognize that the overall bias is not confined to off-policy bias. It also includes over-optimistic bias \citep{hasselt2010double,van2016deep,fujimoto2018addressing} and hindsight biases \citep{blier2021unbiased,schramm2023usher}. Therefore, to isolate the effects and better understand the intricate nature of off-policy bias, we utilize existing techniques \citep{fujimoto2018addressing} to counteract the over-optimistic bias and completely bypass hindsight biases stemming from stochastic dynamics \citep{blier2021unbiased,schramm2023usher} by centering our study on deterministic environments.

Our main contributions are summarized as follows:
i)  By analyzing off-policy bias in multi-step GCRL, we dissect off-policy bias in multi-step GCRL, categorizing it into two distinct types based on their respective roles, and introduce metrics for evaluating each type of bias.
ii) We probe the root causes of two types of biases in multi-step GCRL, elucidating their beneficial and detrimental effects on learning.
iii) We propose a novel resilient strategy for each type of off-policy bias in multi-step GCRL, culminating in the robust algorithm BR-MHER.
iv)  Empirical evaluation demonstrates that our approach robustly outperforms both the HER baseline and several state-of-the-art multi-step GCRL methods in efficiency, bias mitigation, and performance across varied GCRL benchmarks.
\section{Related Work}
\paragraph{Multi-Step Off-Policy RL}
In off-policy multi-step RL, addressing off-policy bias is paramount. There are several identified biases: overestimation bias stemming from function approximation errors \citep{hasselt2010double,van2016deep,fujimoto2018addressing}, distributional shift bias due to imbalanced learning distributions \citep{schaul2015prioritized}, and hindsight bias which manifests in stochastic environments \citep{blier2021unbiased,schramm2023usher}. Both TD(\(\lambda\)) and Q(\(\lambda\)) methods utilize eligibility traces as a means to mitigate such biases \citep{sutton2018reinforcement,harutyunyan2016q}. Additionally, Q(\(\lambda\)) \citep{harutyunyan2016q} brings in off-policy corrections using the current action-value functions. Importance Sampling (IS) offers an additional layer for bias correction and aids in convergence \citep{precup2001off,munos2016safe}, but it does come with a trade-off of heightened variance. Despite these corrective measures, a majority of these methods tend to cap their step size, often limiting it to three steps \citep{barth2018distributed,hessel2018rainbow}.

\paragraph{Multi-Step Off-Policy GCRL}
Multi-step GCRL has inherent biases akin to traditional multi-step RL, significantly affecting learning. The prominence of off-policy bias intensifies in multi-step GCRL, especially as the reward accumulation steps increase, magnifying both bias and variance \citep{yang2021bias}. \cite{yang2021bias} presented MHER, MHER(\(\lambda\)), and MMHER, which merge multi-step learning with hindsight relabeling \citep{andrychowicz2017hindsight}. While vanilla MHER lacks mechanisms to handle off-policy bias, MHER(\(\lambda\)) extends the MHER base by leveraging techniques from TD(\(\lambda\)) \citep{sutton2018reinforcement}. MMHER, meanwhile, uses an on-policy target from a learned world model. The robustness of MHER(\(\lambda\)) and MMHER hinges on MHER's inherent stability and the precision of the learned model, respectively. In contrast, our study introduces two novel subtypes of multi-step off-policy bias in GCRL: \emph{shooting} and \emph{shifting} biases. The latter, unique to GCRL, emerges from accumulating off-policy biases at goal states, influencing prior state value estimates. We found pronounced shifting biases in multi-step GCRL for intricate tasks. Our truncation technique effectively curtails this bias. Our insights suggest specific behavior policies might offer benefits. With this, we deploy quantile regression to address off-policy bias comprehensively.
\paragraph{Quantile-Based Regression}
Quantile regression was pioneered by \cite{koenker2001quantile} and later adapted for distributional RL to model state-action value functions' quantile function \citep{dabney2018implicit,dabney2018distributional,kuznetsov2020controlling}. Expanding on this, \cite{tang2022nature} integrated multi-step RL within distributional RL frameworks utilizing quantile regression. Motivated by these studies, our exploration specifically targets the upper quantiles of the value distribution. While rooted in deterministic settings, we account for stochasticity introduced by multi-step behavior policies, which sets our work apart from existing studies. Unlike prior works learning a complete stochastic value distribution, we prioritize the upper quantile, representing optimal outcomes under select effective behavior policies tied to goals. This quantile guides agents toward actions likely surpassing current estimates and counteracts suboptimal behavior policies' drawbacks.
\section{Preliminary}\label{sect:preliminary}
GCRL is framed as a goal-augmented MDP characterized by parameters $(\mathcal{S}, \mathcal{A}, \mathcal{T}, r, \mathcal{G}, p_{d\!g}, \phi, \gamma, T)$. Here, $\mathcal{S}$, $\mathcal{A}$, $\gamma$, and $T$ are state space, action space, discount factor, and horizon, respectively. The transition function \(\mathcal{T}(s'|s,a)\) gives the probability of transitioning from state \(s\) to \(s'\) via action \(a\). The goal space is $\mathcal{G}$, with $p_{d\!g}$ as the desired goal distribution. The function $\phi(s)$ maps state $s$ to its achieved goal.
In sparse reward settings, the reward \(r(s', g)\) is binary, being zero if the distance \(d(\phi(s'), g)\) between achieved and target goals is less than \(\epsilon\) and \(-1\) otherwise. The success of an episode \(\tau\) with goal \(g\) is defined by \(r(s_T^\tau, g)=0\). It is worth noting that merely reaching the goal does not constitute the termination of an episode. This distinction is crucial as an agent might reach the goal accidentally and lack the ability to sustain its presence there. For instance, if an object slides through the goal position at a non-zero speed, this is not considered a successful outcome. The agent must reach the goal and maintain its position there for the episode to be successfully completed.  GCRL's objective is to learn a goal-conditioned policy $\pi(s, g)$ maximizing:
\begin{equation}
    J(\pi) = \mathbb{E}_{g\sim p_{d\!g}, a_t \sim \pi(s_t, g), s_{t+1}\sim \mathcal{T}(\cdot|s_t, a_t)}\left[\sum_{t=0}^{T-1} \gamma^t r(s_{t+1}, g)\right].  \nonumber
\end{equation}
With an actor-critic framework, the policy (actor) \(\pi_\psi\) and the action-value function (critic) \(Q_\theta\) are parameterized by \(\psi\) and \(\theta\), respectively. The policy \(\pi_\psi\) is updated by minimizing the following loss function:
\begin{equation}
L(\psi) = - \mathbb{E}_{\tau \sim \mathcal{B},\, t \sim [0, T),\, g \sim h(t, \tau)}\left[Q_\theta(s^\tau_t, \pi_\psi(s^\tau_t, g), g)\right],  \label{eq:actor_loss} 
\end{equation}
where \(\tau\) represents a trajectory sampled from the replay buffer \(\mathcal{B}\), \(t\) is a timestep sampled within the horizon \(T\), \(s_t^\tau\) denotes the state at timestep \(t\) for trajectory \(\tau\), and \(h(t, \tau)\) is the distribution of goals according to the hindsight relabeling technique. The action-value function, \(Q_\theta\), is updated by minimizing the following loss function:
\begin{equation}
L(\theta) = \mathbb{E}_{\tau \sim \mathcal{B},\, t \sim [0, T),\, g \sim h(t, \tau)}\left[\left(Q_\theta(s^\tau_t, a^\tau_t, g) - y^{(n)}(t, \tau, g)\right)^2 \right], \label{eq:expected_multi_step_gcrl}
\end{equation}
where \(a^\tau_t\) is the action at the \(t\)th step of trajectory \(\tau\), and \(y^{(n)}(t, \tau, g)\) calculates the \(n\)-step target value, defined as:
\begin{equation}
y^{(n)}(t, \tau, g) = \sum_{i=1}^{n-1}\gamma^i r(s^\tau_{t+i}, g) + \gamma^n Q_{\bar{\theta}}(s^\tau_{t+n}, \pi_{\bar{\psi}}(s^\tau_{t+n}, g), g),
\end{equation}
where \(\bar{\psi}\) and \(\bar{\theta}\) are the parameters of the target actor and critic for learning stabilization.
Although \(n\) can truncate if \(t+n > T\), this is simplified for clarity. Typically, \( \pi \) represents \(\pi_\psi\), with emphasis on learning the action-value function. Further context is provided in Appendix A.

\section{Bias Decomposition} \label{sect:bias_decomposition}
This section delves into the off-policy bias in multi-step GCRL. Through comprehensive analysis, we decompose this bias into two distinct types and examine their implications on action-value estimation.

The temporal difference (TD) error between two consecutive states at the step \( t \) of trajectory \( \tau \) is
\begin{equation}
    \delta_\Theta(s^\tau_t, a^\tau_t, g) =  \gamma Q_{\bar{\theta}}(s^\tau_{t+1}, \pi_{\bar{\psi}}(s^\tau_{t+1}, g), g) + r(s^\tau_{t+1}, g) - Q_\Theta(s^\tau_t, a^\tau_t, g). \nonumber 
\end{equation}
The variable \(\Theta\) in the action-value function \( Q_\Theta \) can take the values \(\theta\) or \(\bar{\theta}\), determining whether the temporal difference (TD) error is computed with respect to the critic or the target critic, respectively. In a deterministic environment, \(\delta_\theta(s_t^\tau, a_t^\tau, g)\) is equivalent to \(\delta_\theta(s, a, g)\) whenever \((s_t^\tau, a_t^\tau, g)\) equals \((s, a, g)\).
The $n$-step TD error between the $n$-step targets and the estimated action-value is characterized as
\begin{equation}
 y^{(n)}(t, \tau, g) - Q_{\theta}(s^\tau_{t}, a^\tau_{t}, g) = \delta_{\theta}(s^\tau_t, a^\tau_t, g) + \sum_{i=1}^{n-1} \gamma^i[A_{\bar{\theta}}(s^\tau_{t+i}, a^\tau_{t+i}, g) + \delta_{\bar{\theta}}(s^\tau_{t+i}, a^\tau_{t+i}, g)], \label{eq:off_policy_n_step_TD}
\end{equation}
 where \( A_{\bar{\theta}}(s^\tau_{t+i}, a^\tau_{t+i}, g) =Q_{\bar{\theta}}(s^\tau_{t+i}, a^\tau_{t+i}, g) - Q_{\bar{\theta}}(s^\tau_{t+i}, \pi(s^\tau_{t+i}, g), g) \) is the estimated advantage of action \( a^\tau_{t+i} \) in state \( s^\tau_{t+i} \), given goal $g$. The details of the derivation of Eq.~(\ref{eq:off_policy_n_step_TD}) can be found in Appendix A.  
Through the application of Eq.~(\ref{eq:off_policy_n_step_TD}), the critic loss in Eq.~(\ref{eq:expected_multi_step_gcrl}) is reformulated as follows:
\begin{equation}
L(\theta) = \mathbb{E}_{\tau,\, t,\, g}\left[ \left( \delta_\theta(s_t^\tau, a_t^\tau, g) - \sum_{i=1}^{n-1} \gamma^i \left( A_{\bar{\theta}}(s^\tau_{t+i}, a^\tau_{t+i}, g) + \delta_{\bar{\theta}}(s^\tau_{t+i}, a^\tau_{t+i}, g) \right) \right)^2 \right],
\label{eq:expected_multi_step_gcrl_delta}
\end{equation}
where the summation term \(\sum_{i=1}^{n-1}\) becomes zero when \(n=1\). In single-step learning ($n=1$), \(\delta_\theta(s_t^\tau, a_t^\tau, g) = 0\) for all \(\tau \in \mathcal{B}\), \(t \in [0, T)\), and \(g\) within the support of goal distribution \(h(\tau, t)\) constitutes the optimal solution for \(L(\theta)\). However, this condition does not necessarily hold as \(n\) increases, owing to the additional summation term. In a deterministic environment, \(\delta_\theta(s_t^\tau, a_t^\tau, g)\) is equivalent to \(\delta_\theta(s, a, g)\) whenever \((s_t^\tau, a_t^\tau, g)\) equals \((s, a, g)\). Nonetheless, the summation \(\sum_{i=1}^{n-1} \gamma^i \left( A_{\bar{\theta}}(s^\tau_{t+i}, a^\tau_{t+i}, g) + \delta_{\bar{\theta}}(s^\tau_{t+i}, a^\tau_{t+i}, g) \right)\) varies with the trajectory \(\tau\) and the timestep \(t\), thereby deviating the optimal TD error \( \delta_\theta(s_t^\tau, a_t^\tau, g) \) for \( L(\theta)\) from zero. \(\delta_\Theta(s, a, g)\) itself is no longer attributed to the network approximation error any more as it is not optimized towards zero. Consequently, in multi-step learning scenarios, \(\delta_\theta(s, a, g)\) encompasses biases originating from both the off-policy advantage and the TD error components, as reflected in the summation term. Given that the parameter \(\bar{\theta}\) of the target critic is updated from the parameter \(\theta\) of the critic, it follows that \(\delta_{\bar{\theta}}(s, a, g)\) is similarly affected by these biases. 
Subsequently, we will explore how this localized bias \( \delta_\theta(s_t^\tau, a_t^\tau, g) \) aggregates to form a broader off-policy bias in the action-value function.

For on-policy evaluation, let \( \tau^g \) denote the trajectory resultant from policy \( \pi(\cdot, g) \). Recursive decomposition of the learned action-value \( Q_{\bar{\theta}}(s^{\tau^g}_t, a^{\tau^g}_t, g) \) and its unbiased counterpart \( Q_\pi(s^{\tau^g}_t, a^{\tau^g}_t, g) \) along \( \tau^g \) until episode termination yields:
\begin{align}
Q_{\bar{\theta}}(s^{\tau^g}_t, a^{\tau^g}_t, g) &= \sum_{i=t}^{T-1} \gamma^{i-t}[r(s^{\tau^g}_{i+1}, g)+ \delta_{\bar{\theta}}(s^{\tau^g}_i, a^{\tau^g}_i, g)] + \gamma^{T-t} Q_{\bar{\theta}}(s^{\tau^g}_T, \pi(s^{\tau^g}_T, g), g), \label{eq:decompsition}\\
Q_\pi(s^{\tau^g}_t, a^{\tau^g}_t, g) &= \sum_{i=t}^{T-1} \gamma^{i-t}r(s^{\tau^g}_{i+1}, g) + \gamma^{T-t} Q_\pi(s^{\tau^g}_T, \pi(s^{\tau^g}_T, g), g). \label{eq:true_decomposition}
\end{align}
In the context of multi-step GCRL, our aim is to devise a strategy that minimizes the detrimental effects of off-policy bias, thereby augmenting the agent's proficiency in achieving goals. Given this focus, off-policy bias in failing policies becomes less of interest given our target. Consequently, our analysis intentionally focuses on successful evaluation trajectories, denoted as \( \tau^{g+} \). Analyzing off-policy bias in successful trajectories, together with overall success rates, provides a multidimensional perspective. This approach not only gives a thorough assessment of the algorithm's performance but also reveals how bias adversely affects learning efficiency and success rates.

In the context of a successful trajectory \( \tau^{g+} \), both the terminal reward \( r(s^{\tau^{g+}}_T,g) \) and the terminal action-value \( Q_\pi(s^{\tau^{g+}}_T, \pi(s^{\tau^{g+}}_T, g), g) \) are zero, in accordance with the success criteria outlined in Section~\ref{sect:preliminary}.
Given these conditions and Eqs.~(\ref{eq:decompsition}) and (\ref{eq:true_decomposition}), we discern the off-policy bias as:
\begin{align}
    B(s^{{\tau}^{g+}}_t, a^{{\tau}^{g+}}_t, g) &= \underbrace{\sum_{i=t}^{T-1} \gamma^{i-t} \delta_{\bar{\theta}}(s^{{\tau}^{g+}}_i, a^{{\tau}^{g+}}_i, g)}_{\text{Shooting Bias}}
    \phantom{=} + \gamma^{T-t} \underbrace{ \vphantom{\sum_{i=t}^{T-1}} Q_{\bar{\theta}}(s^{{\tau}^{g+}}_T, \pi(s^{{\tau}^{g+}}_T, g), g)}_{\text{Shifting Bias}}.
     \label{eq:bias_decomposition}
\end{align}

 Thus, we dissect the off-policy bias in Eq.~(\ref{eq:bias_decomposition}) into two distinct segments: the \textit{shooting bias} and the \textit{shifting bias}. The shooting bias, accumulating within the finite time horizon \( T \), can distort the action-value function and affect decision-making. The shifting bias manifests mainly within goal states which accumulate over paths of potentially infinite lengths, , detailed further in Section~\ref{sect:shifting_bias}. Both biases can be fundamentally decomposed into a discounted sum of TD errors, which tend to deviate from zero due to the influence of off-policy data. This decomposition becomes particularly evident when learning with the loss function described in Eq.~(\ref{eq:expected_multi_step_gcrl_delta}), especially for cases where \(n > 1\).
 While the shooting bias in multi-step GCRL shows similarities to off-policy biases in traditional multi-step RL, the shifting bias is unique to multi-step GCRL.

\section{Bias Management and Measurement}

In this section, we analyze the shooting and shift biases, propose strategies to manage them and introduce metrics to quantify them.

\subsection{Shooting Bias Management}
A direct approach to mitigate the shooting bias is importance sampling (IS) \citep{precup2001off}, aiming to correct the off-policy bias. However, IS may suffer from the variance problem. Moreover, IS may not effectively harness multi-step targets to enhance learning efficiency. For deriving unbiased targets via importance sampling, the effects of the data in the multi-step transitions reduce as the step increases, particularly evident in policies exhibiting low entropy. In extreme scenarios, multi-step learning for a deterministic policy could revert to single-step learning. Though it appears reasonable to disdain the off-policy data less irrelevant to the current policy, we argue that it is worth rethinking a pivotal question:  \textit{Is bias always detrimental, or can specific biases positively guide and improve the target policy?}

\begin{figure}
    \centering
    \includegraphics[width=\textwidth]{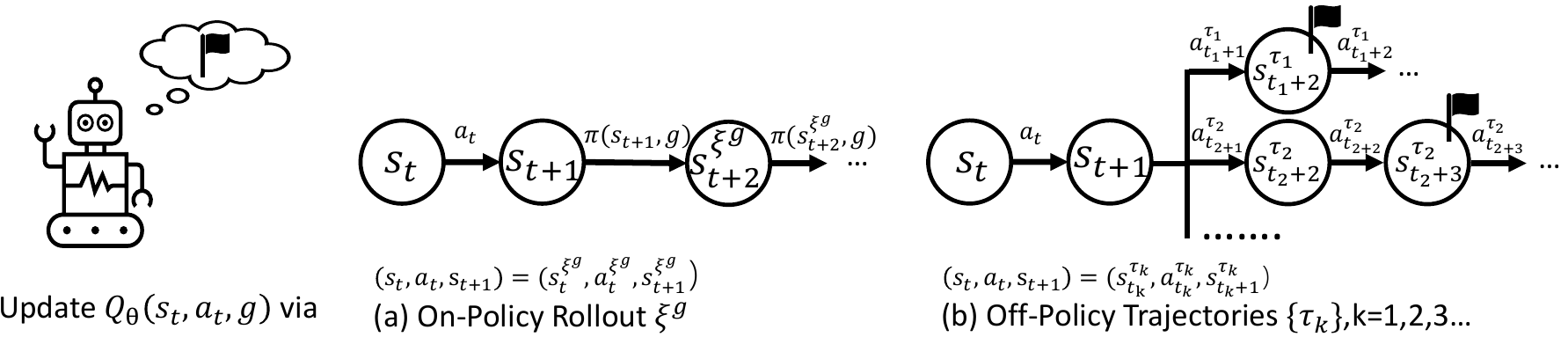}
    \caption{Illustration of \( Q_\theta(s_t, a_t, g) \) updates in a deterministic environment using two approaches. (a) On-policy rollout \( \xi^g \), where the agent learns from transitions following the sequence \( (s_t, a_t, s_{t+1}) \). (b) Off-policy trajectories \( \{\tau_k\} \), with \( k = 1, 2, \cdots \), where the agent learns from multi-step transitions in each trajectory \( \tau_k \) at timestep \( t_k \), characterized by \( (s_t, a_t, s_{t+1}) = (s^{\tau_k}_{t_k}, a^{\tau_k}_{t_k}, s^{\tau_k}_{t_k+1}) \)}
    \label{fig:beneficial_bias}
\end{figure}

To explore this, we examine the benefits of multi-step learning using off-policy trajectories compared to on-policy rollouts.  For clarity in multi-step indexing, we denote the state-action pair \( (s, a) \) with a subscript \( t \), as \( (s_t, a_t) \). In a deterministic environment, the subsequent state \( s_{t+1} \) is deterministically derived from \( (s_t, a_t) \). The on-policy rollout originating from the state-action pair \( (s_t, a_t) \) is denoted as \( \xi^g \). For the set of off-policy trajectories that include \( (s_t, a_t) \), each trajectory is represented as \( \tau_k \), where \( k \) serves as the index. The timestep in trajectory \( \tau_k \) corresponding to \( (s_t, a_t) \) is denoted as \( t_k \).
 As depicted in Fig.~\ref{fig:beneficial_bias}(a), the on-policy rollout \( \xi^g \) can entirely avoid bias relative to the current policy. In contrast, as illustrated in Fig.~\ref{fig:beneficial_bias}(b), off-policy trajectories introduce an off-policy bias by computing target values along these trajectories.

When an agent derives a higher target value \( y^{(n)}(t_k, \tau_k, g) \) from an off-policy trajectory \( \tau_k \) compared to \( y^{(n)}(t, \xi^g, g) \) obtained from the corresponding on-policy rollout \( \xi^g \), this suggests that the actions in \( \tau_k \) starting from timestep \( t_k \) might offer better strategies over the next \( n \) steps. For instance, as illustrated in Fig.~\ref{fig:beneficial_bias}(b), trajectory \( \tau_1 \) achieves the goal in fewer steps from state \( s_t \) compared to the on-policy rollout depicted in Fig.~\ref{fig:beneficial_bias}(a). This scenario, particularly in two-step learning for updating \( Q_\theta(s_t,a_t,g) \), implies that \( y^{(2)}(t_k, \tau_k, g) > y^{(2)}(t, \xi^g, g) \) may indicate the superiority of the action \( a^{\tau_1}_{t_1+1} \) over \( \pi(s_{t+1}, g) \). Consequently, \( y^{(2)}(t_1, \tau_1, g) \) provides a more accurate representation of the potential benefits of taking action \( a_t \) in contrast to \( y^{(2)}(t, \xi^g, g) \).

Relying exclusively on on-policy rollouts for updating \( Q_\theta(s_t,a_t,g) \) can lead to inefficiencies, particularly in mini-batch learning scenarios where the agent must wait to recognize and incorporate the advantages of actions like \( a_{t_1+1}^{\tau_1} \) in the subsequent steps. This delayed policy adjustment process may hinder the agent's ability to promptly leverage more beneficial actions such as \( a_{t} \). To effectively utilize advantageous off-policy strategies, we introduce the concept of an \( n \)-step off-policy transition that exhibits beneficial off-policy bias.

\begin{definition}
    In deterministic environments, for a given starting state-action pair \( (s_t, a_t) \), an \( n \)-step transition sequence \( s^{\tau_k}_{t_k}, a^{\tau_k}_{t_k}, \cdots, s^{\tau_k}_{t_k+n} \) in trajectory \( \tau_k \), where \( (s_t, a_t) = (s^{\tau_k}_{t_k}, a^{\tau_k}_{t_k}) \), is said to exhibit \textit{beneficial off-policy bias} when the \( n \)-step target values conditioned on the goal \( g \), derived from this off-policy transition sequence, are greater than those obtained from the corresponding \( n \)-step on-policy rollout \( \xi^g \). This property is mathematically expressed as follows.
    \begin{equation}
      y^{(n)}(t_k, \tau_k, g) > y^{(n)}(t, \xi^g, g).
    \end{equation}
\end{definition}
 However, detrimental off-policy bias co-exists with beneficial bias. Some targets even approach the worst target value to be taken $\frac{-1}{1-\gamma}$, making it extremely hard to compare the potential of each action. Thus, there arises another problem: \textit{Can we make use of the beneficial bias to improve the learning efficiency while mitigating the detrimental effects of off-policy bias to the learning process?}

In the model-free setting, where we often lack access to on-policy rollouts \( \xi_g \), accurately identifying beneficial bias presents a significant challenge. Recognizing that higher target values in multi-step transitions typically stem from more effective strategies or decisions, we infer that these transitions are likely to exhibit higher beneficial bias or lower detrimental bias. This inference guides our focus on such transitions during the learning process. Accordingly, we define the maximum target value \( y_{\text{max}}(s, a, g) \) and the corresponding learning loss \( L_{\text{max}}(\theta) \) as:
\begin{align}
    y_{\text{max}}(s, a, g) &= \max_{t_k, \tau_k} \left\{ y^{(n)}(t_k, \tau_k, g) \mid s_t^{\tau_k} = s, a_t^{\tau_k} = a \right\}, \nonumber\\
    L_{\text{max}}(\theta) &= \mathbb{E}_{\tau, t, g}\left[\left(Q_\theta(s^\tau_t, a^\tau_t, g) - y^{(n)}(t, \tau, g)\right)^2 \big | y^{(n)}(t, \tau, g) = y_{\text{max}}(s^\tau_t, a^\tau_t, g) \right]. \nonumber
\end{align}
Though obtaining \( y_{\text{max}}(s, a, g) \) is feasible in a tabular setting due to its limited state-action space, this task becomes significantly more challenging in scenarios with high-dimensional state-action space. An alternative method involves learning only towards target values larger than the current action-value estimates, guiding the agent iteratively towards higher values until the maximum is reached. This is formalized in the loss function \( L_{\text{larger}}(\theta) \):
\begin{equation}
    L_{\text{larger}}(\theta) = \mathbb{E}_{\tau, t, g}\left[\left(Q_\theta(s^\tau_t, a^\tau_t, g) - y^{(n)}(t, \tau, g)\right)^2 \big | y^{(n)}(t, \tau, g) > Q_\theta(s^\tau_t, a^\tau_t, g) \right]. \nonumber
\end{equation}
However, the approach of consistently learning towards larger target values can inadvertently foster over-optimization bias in the action-value function, as it relentlessly drives the learning process to seek higher values. This unidirectional focus might impede the action-value function's adaptability to fluctuations in target value distributions. Notably, action values often start near zero following neural network initialization and can decrease significantly during learning, potentially skewing the identification of the true maximum value.

\paragraph{Bias Management with Quantile Regression}\label{sect:quantile_regression}
Rather than pursuing the maximum, we reorient our focus towards learning the action-value function to capture an upper quantile \( \rho \) (\( 0.5 < \rho < 1\)) of the target values. To learn the upper quantile, we employ quantile regression \citep{koenker2001quantile,dabney2018distributional}. In our methodology, we adopt the loss function \( l_{QR} \) from the quantile regression framework to quantify the discrepancy between each action-value \( Q \) and the corresponding target action-value \( y \). Incorporating both the quantile level \( \rho \) and the Huber loss threshold \( \kappa \) \citep{huber1992robust}, the formulation of \( l_{QR} \) is:
\begin{equation}
l_{QR}(y, Q; \rho, \kappa) = \left| \rho - \mathbb{I}(y, Q) \right| \cdot \begin{cases}
(y - Q)^2 & \text{if } \left|y - Q\right| \leq \kappa, \\
\kappa \left( 2 \cdot \left|y - Q\right| - \kappa \right) & \text{otherwise}.
\end{cases} \label{eq:quantile_loss}
\end{equation}
where the indicator function \( \mathbb{I}(y, Q) \) equals 1 if \( y < Q \) and 0 otherwise. Instead of simply taking the average, $l_{QR}$ weight target values higher and lower than the current action-value by $|\rho|$ than $|1-\rho|$, respectively. By applying quantile regression with the \( l_{QR} \) loss function, we can reformulate the general loss in Eq.~(\ref{eq:expected_multi_step_gcrl}) into this expression:
\begin{equation}
L_{QR}(\theta) = \mathbb{E}_{\tau, t, g}\left[l_{QR}(y^{(n)}(t, \tau, g), Q_\theta(s^\tau_t, a^\tau_t, g); \rho, \kappa)\right]. \label{eq:quantile_regression}
\end{equation} 
This method empowers the agent to more effectively exploit beneficial biases while simultaneously mitigating the impact of detrimental biases. Furthermore, it fosters adaptability to variations in the target action-value distribution, thereby maintaining consistent responsiveness to the dynamics of action-value changes.

\subsection{Shifting Bias Reduction}\label{sect:shifting_bias}

Contrasting with shooting bias, which is confined within a finite temporal horizon $T$ as illustrated in Eq.~(\ref{eq:bias_decomposition}), shifting bias distinctively accumulates in goal states over an infinite horizon. This bias escalates as learning advances, leading to a consequential shift in the action-value of the goal state. Such a shift tends toward a biased estimation, diverging from the actual action-value and thus significantly altering the learning trajectory. In GCRL, the update mechanism for state-action pairs, preceding the achievement of the goal, is fundamentally reliant on the action-value at the goal state. Therefore, a shift in the goal state's action-value propagates backwards, affecting the entire chain of state-action pairs.  While quantile regression, as discussed previously, offers a viable approach to mitigate these biases and guide the agent towards a more robust off-policy strategy, the continual shift of target action-values poses a challenge to learning stability. It underscores the necessity of developing strategies that can adaptively respond to and compensate for these biases. 

For the shifting bias, as detailed in Eq.~(\ref{eq:true_decomposition}), extending the episode's unrolling to the infinite future from time \( T \) allows for its decomposition into:
\begin{equation}
     Q_{\bar{\theta}}(s^{{\tau}^{g+}}_T, \pi(s^{{\tau}^{g+}}_T, g), g) = \sum_{i=T}^\infty \gamma^{i-T}[r(s^{\tau^{g+}}_{i+1}, g)+ \delta_{\bar{\theta}}(s^{\tau^{g+}}_i, a^{\tau^{g+}}_i, g)] \label{eq:inf_expansion}
\end{equation}
Given that \( \tau^{g+} \) is a successful trajectory, we assume that for all subsequent states following the policy, the rewards satisfy \( r(s^{\tau^{g+}}_{i+1}, g)=0 \) for all \( i \geq T \). This assumption allows us to simplify Eq.~(\ref{eq:inf_expansion}) as follows:
\begin{equation}
      Q_{\bar{\theta}}(s^{{\tau}^{g+}}_T, \pi(s^{{\tau}^{g+}}_T, g), g) = \sum_{i=T}^\infty \gamma^{i-T}\delta_{\bar{\theta}}(s^{\tau^{g+}}_i, a^{\tau^{g+}}_i, g)
     \label{eq:inf_expansion_simp}
\end{equation}
Unlike shooting bias, which is inherently constrained by the episode horizon and generally controllable, shifting bias accumulates over an infinite horizon. As outlined in Sect. \ref{sect:bias_decomposition}, while TD errors \( \delta_{\bar{\theta}} \) are typically minimized towards zero in single-step learning, they often deviate from this baseline due to the influence of off-policy bias in multi-step scenarios. Such deviations can be exacerbated by self-loops during iterated learning, potentially escalating even minor TD errors into significant shifting biases over the infinite horizon. Consequently, for transitions among goal states, we prefer reverting to single-step learning to minimize TD errors in these specific transitions. Simultaneously, we aim to retain the benefits of multi-step learning for states preceding the goal states.

\paragraph{Truncated Multi-Step Targets}
To counteract the shifting bias, we introduce the strategy of truncating multi-step transitions at the first goal state. This approach reverts transitions among goal states to single-step learning, effectively minimizing TD errors in these segments. Meanwhile, for transitions that do not reach the goal state, we maintain the benefits of multi-step learning. The truncated multi-step target, denoted by \( \hat{y}^{(n)}\), is computed as follows:
\begin{equation}
\hat{y}^{(n)}(t, \tau, g) = r(s^{\tau}_{t+1}, g) + \begin{cases}
 \gamma \hat{y}^{(n-1)}(t+1, \tau, g) & \text{if } r(s^{\tau}_{t+1}, g) \neq 0 \text{ and } n>1, \\
\gamma Q_\btheta(s_{t+1}^\tau, \pi(s_{t+1}^\tau, g), g)& \text{otherwise}.
\end{cases} \label{eq:tmher}
\end{equation}
This method effectively reduces shifting bias, thereby ensuring stable learning targets for preceding state-action pairs. As will be demonstrated in our subsequent experimental results, this approach contributes significantly to the stability and efficacy of the learning process.

\subsection{Bias Resilient Algorithm}
Building upon the technical solutions detailed in previous sections, we have developed the \textit{Bias-Resilient MHER} (BR-MHER) algorithm, which is specifically designed to effectively mitigate both shooting and shifting biases in multi-step GCRL. To enhance the bias mitigation capabilities of BR-MHER, we have integrated the MHER(\(\lambda\)) technique \citep{yang2021bias}. This addition allows for the interpolation of multi-step targets across various time steps, further refining the target value derivation process.

With the integration of MHER(\(\lambda\)), BR-MHER employs the truncated multi-step target \(\hat{y}^{(n)}\) as described in Eq.~(\ref{eq:tmher}), and calculates the new target value as follows:
\begin{equation}
    \hat{y}^{(n)}_\lambda(t, \tau, g) = \frac{\sum_{i=1}^n\lambda^i \hat{y}^{(i)}(t, \tau, g)}{\sum_{i=1}^n\lambda^i}. \label{eq:truncated_mher_lambda}
\end{equation}

A comprehensive outline of the BR-MHER algorithm, incorporating these strategies, is thoughtfully provided in Algorithm \ref{alg:br-mher} for a clear understanding and ease of implementation.

\begin{algorithm}
\caption{Bias Resilient MHER (BR-MHER)}
\begin{algorithmic}[1]
\Require Environment $E$, Replay buffer $\mathcal{B}$, mini-batch size $b$, reward function $r$, policy $\pi_\psi$, quantile $\rho$, Huber loss threshold $\kappa$, action-value functions $Q_{\theta}^\rho$, hindsight relabeling function $h$.
\For{$episode = 1, \dots, N$}
    \State Initialize environment $E$ to obtain initial state $s_0$ and goal $g$.
    \For{$t = 0, \dots, T-1$}
        \State Execute action $a_t \sim \pi(s_t, g)$ and observe the resultant state $s_{t+1}$.
        \State Store transition $(s_t, a_t, s_{t+1}, g)$ in $\mathcal{B}$.
        \State Update state to $s_t \gets s_{t+1}$.
    \EndFor
    \State Sample a batch $B = \{(s_{t_i}^{\tau_i}, a_{t_i}^{\tau_i}, s_{t_i+n}^{\tau_i}, g_{\tau_i})\}_{i=1}^b$ from $\mathcal{B}$.
    \State Apply hindsight relabeling with $\hat{g}_{\tau_i} \sim h(t_i, \tau_i)$ and set $g_{\tau_i} \leftarrow \hat{g}_{\tau_i}$.
    \State Compute the target $\hat{y}_\lambda^{(n)}(t_i, \tau_i, g_{\tau_i})$ using Eq.~(\ref{eq:truncated_mher_lambda}).
    \State Optimize the expected critic loss \( L_{QR}(\theta) \) according to Eq.~(\ref{eq:quantile_regression}) using batch $B$ to update \( \theta \).
    \State Optimize the expected actor loss $L(\psi)$ as per Eq.~(\ref{eq:actor_loss}) using batch $B$ to update $\psi$.
\EndFor
\end{algorithmic}\label{alg:br-mher}
\end{algorithm}

 \subsection{Bias Measurement} \label{sect:measurement_metrics}
 In analyzing off-policy bias, we aim to quantify shooting and shifting biases as in Eq.~(\ref{eq:bias_decomposition}), introducing metrics \textit{Terminal Shifting Bias} (TSB) and \textit{Initial Shooting Bias} (ISB) defined as follows:
\begin{align}
TSB &= \mathbb{E}_{g,\tau^{g+}} \left[ Q_\theta(s^{\tau^{g+}}_T, a^{\tau^{g+}}_T, g) \right], \label{eq:TSB} \\
ISB &= \mathbb{E}_{g,\tau^{g+}} \left[ Q_\theta(s^{\tau^{g+}}_0, a^{\tau^{g+}}_0, g) - \sum_{i=0}^{T-1} \gamma^{i} r(s_{i+1}^{\tau^{g+}}, g) \right] - \gamma^T\cdot TSB. \label{eq:ISB}
\end{align}
In the metrics, TSB is calculated as the average shifting bias of the sampled on-policy evaluation trajectories. Furthermore, ISB is designed to capture the shooting bias $\sum_{i=0}^{T-1} \delta_\btheta(s_i^{\tau^{g+}}, a_i^{\tau^{g+}}, g)$ which is accumulated over the entire episode horizon. According to Eq.~(\ref{eq:decompsition}), we calculate ISB by subtracting the return and the discounted shifting bias from the action-value of the initial state-action pair, leading to Eq.~(\ref{eq:ISB}). Contrasting with the approach of averaging shooting biases across all state-action pairs, ISB effectively mitigates the diluting effects of such averaging and offers a more precise reflection of the overall severity of shooting bias in the action-value function.

\section{Experiments}
In this section, we evaluate the performance of our BR-MHER algorithm.  The evaluation metrics for the algorithm include sampling efficiency, success rates and off-policy biases. The performance of our approaches is further exemplified through a comparative investigation.
In our experiments, we use six robotic tasks involving three distinct robotic agents: i) Fetch robotic arm \citep{plappert2018multi}: \texttt{FetchPickAndPlace} and \texttt{FetchSlide}; ii) Anthropomorphic robotic hand \citep{plappert2018multi}: \texttt{HandReach} and \texttt{HandManipulateBlockRotateXYZ}; iii) Sawyer robotic arm \citep{nair2018visual}: \texttt{SawyerPushAndReachEnvEasy} and \texttt{SawyerReachXYZEnv}. We also apply our methods in the grid-world tasks \texttt{Simple-MiniGrid-Empty} \citep{gym_minigrid} of scales 25$\times$25 and 50$\times$50. For more details about these tasks, see Appendix~B

\subsection{Baselines}
For a comparative study, we include six experimental baselines: i) HER \citep{andrychowicz2017hindsight}: A straightforward GCRL technique using one-step targets; ii) MHER \citep{yang2021bias}: An expanded version of HER learning through multi-step targets; iii) MHER($\lambda$) \citep{yang2021bias}: A balanced method similar to TD($\lambda$) \citep{sutton2018reinforcement}, harmonizing multi-step targets at various steps via the $\lambda$ parameter; iv) MMHER \citep{yang2021bias}: A model-based approach, calculating on-policy multi-step targets with a world model; v) IS-MHER: A integration of MHER and importance sampling method Retrace($\lambda$)  \citep{munos2016safe}; vi) WGCSL \citep{yang2022rethinking}: An advanced goal-conditioned supervised learning algorithm with meticulously designed weighting mechanisms.

\subsection{Experimental Settings and Implementation} \label{sec:exp_setting}
\paragraph{Experimental Settings} Our experiments aim to answer the following research questions: \textbf{Q1)} How effectively does our BR-MHER manage off-policy bias in multi-step GCRL? \textbf{Q2)} How do our bias mitigation techniques impact learning efficiency? \textbf{Q3)} Does BR-MHER maintain competitive performance with state-of-the-art methods such as IS-MHER and WGCSL in terms of success rate and other relevant metrics? \textbf{Q4)} How do quantile regression and truncated multi-step targets individually and in combination contribute to mitigating off-policy bias in BR-MHER?

We conduct comprehensive experiments, benchmarking our BR-MHER against the baselines across a variety of robotic tasks with $n=3, 5, 7$, and grid-world tasks with $n=3,5,10$. To ensure the reliability of our results, we perform five trials with distinct seeds for each task. Training is stopped after 50 epochs, and the performance is assessed based on the success rate along with the bias and variance metrics, as elaborated in Section~\ref{sect:measurement_metrics}.

\paragraph{Implementation} We utilize a modified OpenAI Baselines codebase \citep{baselines}, following the recommendations made by Yang et al. \citep{yang2021bias}. To mitigate over-estimation bias, we adopt the Clipped Double Q-learning (CDQ) and delayed policy updates from TD3 \citep{fujimoto2018addressing} for all GCRL methods. Our implementation utilizes Multilayer Perceptrons (MLPs) for both actor and critic networks. For a balanced comparison with quantile regression, we used Huber loss for all GCRL methods. The upper quantile ($\rho$) is fixed at 0.75 to balance capturing the upper quantile and adjusting target value distribution shifts. The Huber loss threshold ($\kappa$) is set at 10. 
Detailed specifications of these hyper-parameters and distinct network configurations for various tasks are in Appendix B.
\begin{figure}[ht]
    \centering
    \begin{subfigure}[b]{\textwidth}
        \centering
        \includegraphics[width=\textwidth]{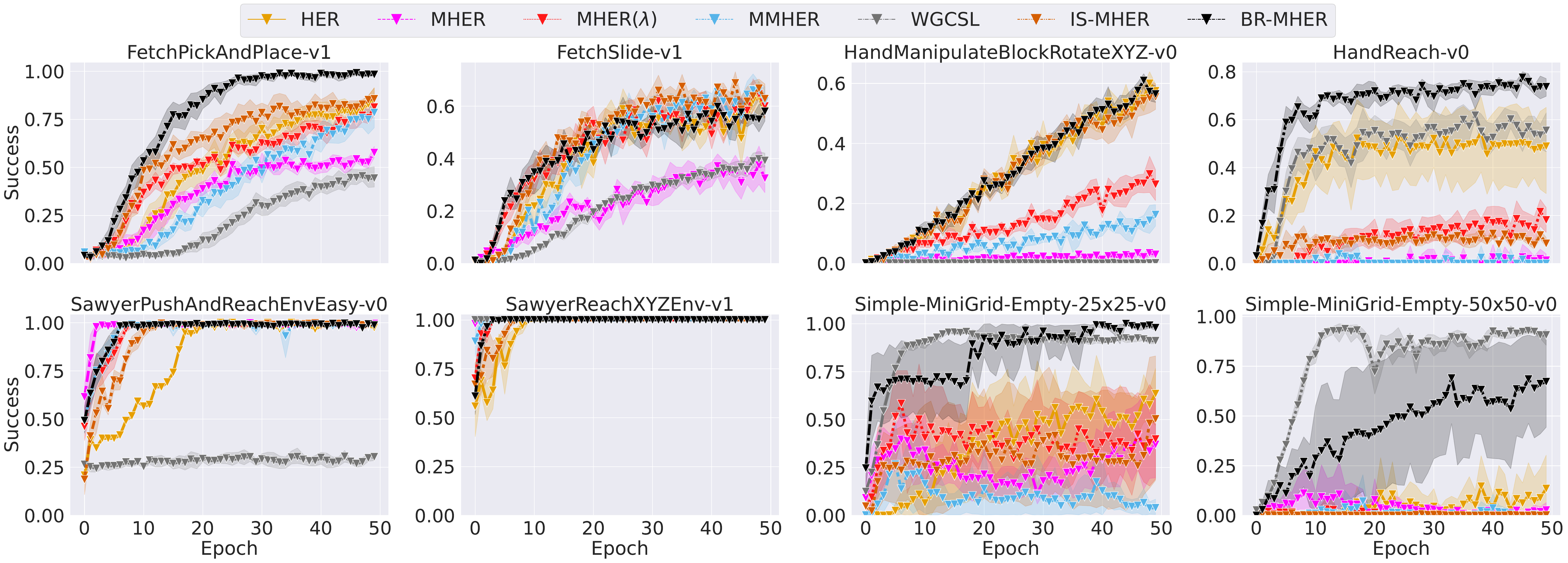}
        \caption{}
        \label{fig:bvr_mher_n7_success}
    \end{subfigure}
    \begin{subfigure}[b]{\textwidth}
        \centering
        \includegraphics[width=\textwidth]{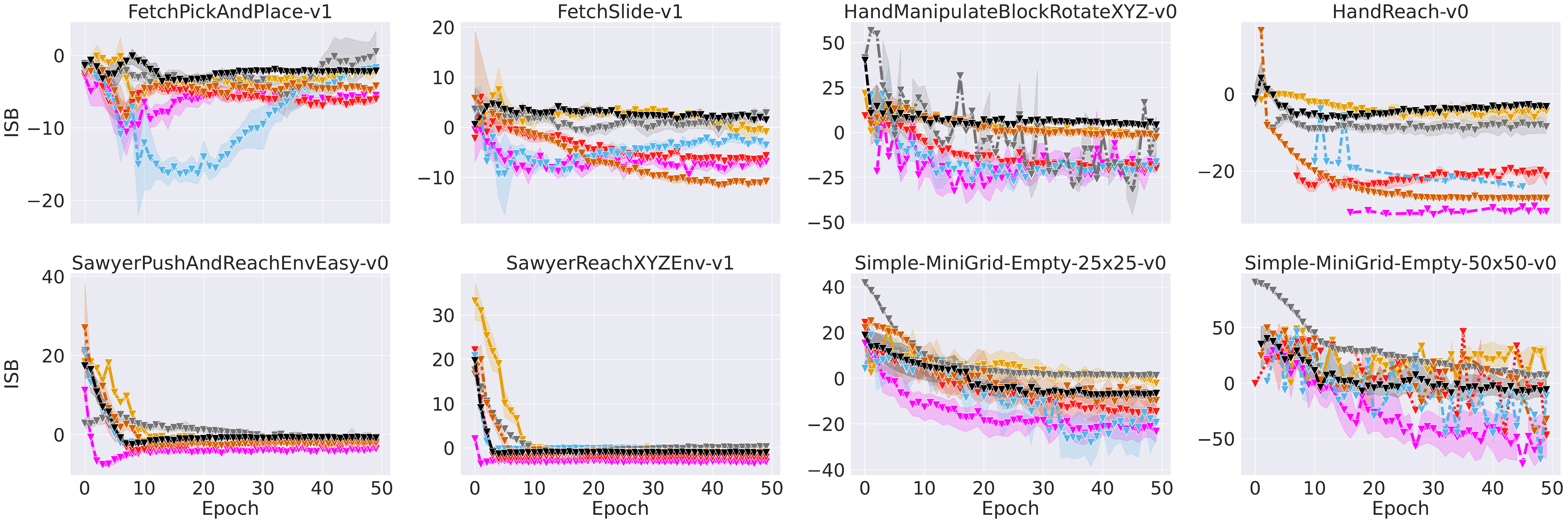}
        \caption{}
        \label{fig:bvr_mher_n7_shooting_bias}
    \end{subfigure}
    \begin{subfigure}[b]{\textwidth}
        \centering
        \includegraphics[width=\textwidth]{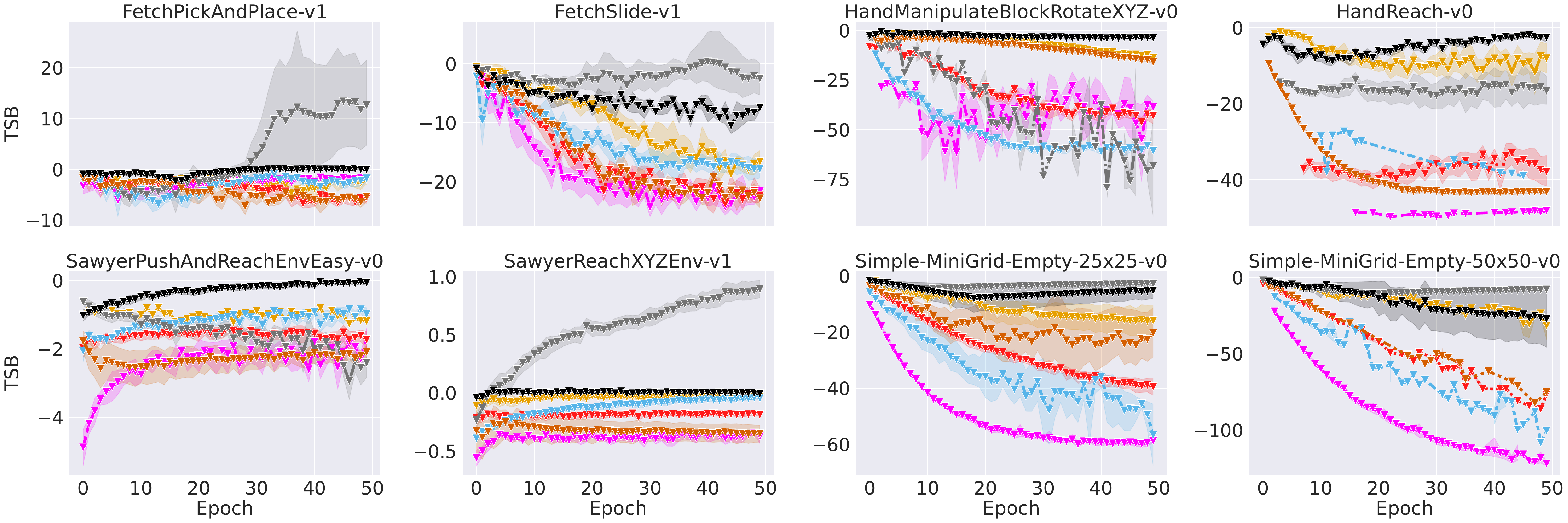}
        \caption{}
        \label{fig:bvr_mher_n7_shifting_bias}
    \end{subfigure}
    \caption{Comparative study of seven-step ($n$=7) GCRL methods on robotic tasks and ten-step ($n$=10) GCRL methods on grid-world tasks. (a) Success rate. (b) ISB for measuring shooting bias. (c) TSB for measuring shifting bias.}
    \label{fig:bvr_mher_n7}
\end{figure}
\subsection{Experiment Results} \label{sect:exp_result}
Due to the limited space, we report only the statistical results (mean and standard deviation) across five random seeds for the multi-step GCRL methods with the largest $n$, specifically $7$ for robotic tasks and $10$ for grid-world tasks. Additional results are in Appendix C.

\paragraph{Comparison with HER} As illustrated in Fig.~\ref{fig:bvr_mher_n7}(a), BR-MHER achieves higher success rates than HER in four tasks and equivalent performance in the remaining tasks. Additionally, BR-MHER outperforms HER in learning efficiency on six tasks and matches efficiency in both \texttt{FetchSlide-v1} and \texttt{HandManipulateBlockRotateXYZ-v0} tasks. Our method attains a shooting bias comparable to HER, as denoted by ISB in Fig.~\ref{fig:bvr_mher_n7}(b). Regarding shifting bias, highlighted by TSB in Fig.~\ref{fig:bvr_mher_n7}(c), BR-MHER records smaller values in six tasks and analogous values in the other two, signifying that even unbiased single-step learning may accumulate errors, detectable by TSB, over an extensive horizon. It also demonstrates that our method effectively mitigates such intrinsic bias in GCRL, alongside off-policy shifting bias.

\paragraph{Comparison with MHER} MHER, a baseline multi-step GCRL method, exhibits high sensitivity to off-policy bias. Fig.~\ref{fig:bvr_mher_n7}(a) demonstrates the superior performance of BR-MHER in success rates and learning efficiency across six tasks. However, in cases where both methods achieve similar success rates, MHER holds a slight edge in learning efficiency, exclusively in the Sawyer tasks.  A detailed bias analysis further highlights consistently larger off-policy bias magnitudes in all tasks by MHER, corroborated by the ISB and TSB metrics in Fig.~\ref{fig:bvr_mher_n7}(b) and \ref{fig:bvr_mher_n7}(c). This bias issue, however, is least severe in the Sawyer tasks, allowing MHER to maintain decent performance.

\paragraph{Comparison with MMHER} MMHER is model-based, in contrast to the model-free approach of BR-MHER.  Fig.~\ref{fig:bvr_mher_n7}(a) shows that BR-MHER generally outperforms or matches MMHER across various tasks, except in \texttt{FetchSlide-v1}.  This task's unique nature, where actions have a limited impact once an object is slid, allows future states to predominantly dictate the return. The world model can swiftly learn these types of transitions, granting MMHER a notable learning advantage. Despite this, BR-MHER demonstrates reduced magnitudes of TSB and ISB across all tasks, indicative of lower off-policy biases as seen in Fig.~\ref{fig:bvr_mher_n7}(b) and \ref{fig:bvr_mher_n7}(c). While MMHER employs on-policy rollouts to avoid these biases, it contends with accumulating model-prediction errors, evidenced in the Fetch tasks in Fig.~\ref{fig:bvr_mher_n7}(b), where an initial increase in ISB possibly reflects heightened model-prediction errors before decreasing as the model better adapts to the environments.

\paragraph{Comparison with MHER($\lambda$)} BR-MHER achieves superior success rates and learning efficiency across five diverse tasks and maintains comparable performance on the Sawyer tasks, as demonstrated in Fig.~\ref{fig:bvr_mher_n7}(a). Despite a slight disadvantage in the \texttt{FetchSlide-v1} task, the ablation study in Sect.~\ref{sect:ablation_study} reveals that the degradation occurs with the introduction of quantile regression. Despite significantly mitigating TSB bias, it poses a challenge in this specific task.
In \texttt{FetchSlide-v1}, a state within the target object's sliding process directly determines future rewards since actions can no longer influence the target object, leading to a largely deterministic distribution of target values. This scenario causes the quantile regression to prefer positive network approximation errors in the target value, culminating in a slightly optimistic shooting bias, as evident by the ISB shown in Fig.~\ref{fig:bvr_mher_n7}(b), and likely hindering learning. Despite this, Figs.~\ref{fig:bvr_mher_n7}(b) and \ref{fig:bvr_mher_n7}(c) confirm that BR-MHER significantly lowers both ISB and TSB, highlighting its ability to mitigate off-policy biases.

\paragraph{Comparison with IS-MHER} BR-MHER outperforms IS-MHER in learning efficiency and success rates across six tasks, showing similar performance in \texttt{HandManipulateBlockRotateXYZ-v0}, but slightly inferior in \texttt{FetchSlide-v1}, as shown in Fig.~\ref{fig:bvr_mher_n7}(a). IS-MHER exhibits larger TSB and ISB magnitudes, outlined in Figs.\ref{fig:bvr_mher_n7}(b) and \ref{fig:bvr_mher_n7}(c). This trend is partly due to the evaluation of a deterministic policy, derived from the stochastic policy used in importance sampling, focused on optimizing success rate evaluations. The disparity is further exacerbated by the use of truncated importance ratios in Retrace($\lambda$) \citep{munos2016safe}.
\paragraph{Comparison with WGCSL}Our comparative assessment highlights a notable inconsistency in WGCSL’s performance across various tasks. Despite exhibiting enhanced learning efficiency in grid-world tasks, as depicted in Fig.~\ref{fig:bvr_mher_n7}(a), it falls behind BR-MHER in success rates for five robotic tasks. As observed in Fig.~\ref{fig:bvr_mher_n7}(c), BR-MHER holds smaller magnitudes of ISB, indicating less severe shooting bias compared to IS-MHER. Although WGCSL employs an action-value function learned via single-step learning, it still exhibits significant shifting biases in certain tasks. This is evidenced by the elevated TSB in \texttt{FetchPickAndPlace-v1} and \texttt{HandManipulateBlockRotateXYZ-v0} (Fig.~\ref{fig:bvr_mher_n7}(c)), correlating with its subpar performance. This discrepancy stems from the supervised policy’s insufficient synchronization with the action-value function's directives, inducing substantial policy and target value fluctuations during the learning process. In contrast, our GCRL methods mitigate this issue, demonstrating reduced TSB in robotic tasks. In grid-world tasks, WGCSL displays markedly lower TSB, indicating diminished fluctuations. In \texttt{Simple-MiniGrid-Empty-50x50}, BR-MHER encounters notable TSB variation among different seeds, leading to a stark success rate oscillation between over 95\% and approximately 30\%. This variation might be attributed to the imbalance between learning action-values in goal states with low variance of target values and states preceding the goals with higher target value variance. This imbalance occasionally prevents the effective minimization of TD errors in goal states, leading to a high magnitude of shifting bias. This observed variation in success rates and the shifting bias underscores BR-MHER’s limitations in handling shifting bias effectively in this specific scenario.

\noindent In response to the first three research questions: \textbf{A1)} BR-MHER substantially minimizes both shooting and shifting bias compared to MHER($\lambda$) and other GCRL methods, as evidenced by the ISB and TSB metrics. \textbf{A2)} BR-MHER generally excels over MHER($\lambda$), achieving enhanced success rates and learning efficiency across multiple tasks, albeit with a slight setback in the \texttt{FetchSlide-v1} task, potentially attributed to over-optimistic shooting bias induced by quantile regression. \textbf{A3)} While IS-MHER and WGCSL exhibit superior performance in specific tasks (\texttt{FetchSlide-v1} and two grid-world tasks respectively), BR-MHER consistently delivers solid and reliable performance across all other tasks and experiments, highlighting its robust and adaptable nature in various contexts.

\subsection{Ablation Study} \label{sect:ablation_study}

This ablation study is systematically designed to unravel the unique contributions of each component embedded within the BR-MHER algorithm. It chiefly emphasizes their roles in the reduction of bias and enhancement of learning efficiency.

As delineated in the primary text, quantile regression and truncated multi-step targets emerge as pivotal techniques, seamlessly integrated within the BR-MHER algorithm. To discern the solitary impact of each, the study explores two algorithmic variants: QR-MHER and TMHER($\lambda$). QR-MHER singularly integrates quantile regression, while TMHER($\lambda$) exclusively embodies truncated multi-step targets, each within the foundational MHER($\lambda$) framework. This methodical investigation aims to shed light on the nuanced influences each component imparts on the overarching performance of the BR-MHER algorithm.

\begin{figure}[t]
    \centering
    \begin{subfigure}[b]{\textwidth}
        \centering
        \includegraphics[width=\textwidth]{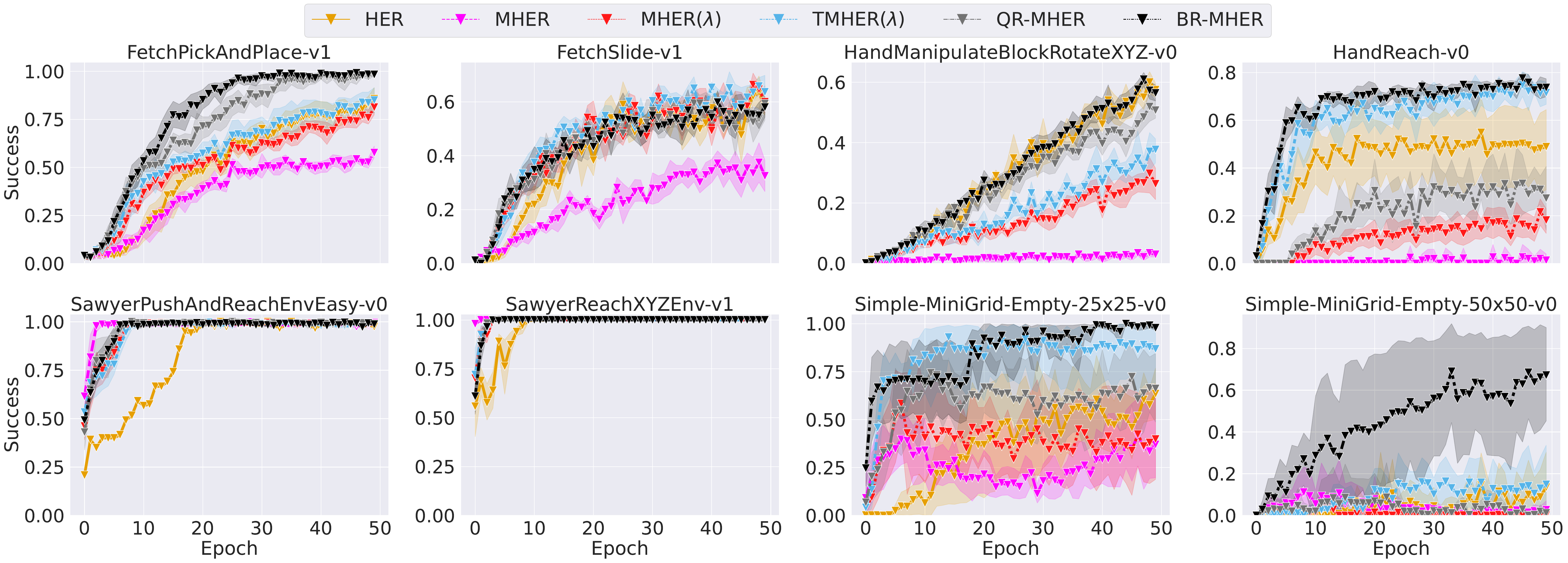}
        \caption{}
        \label{fig:ablation_n7_success}
    \end{subfigure}
    \begin{subfigure}[b]{\textwidth}
        \centering
        \includegraphics[width=\textwidth]{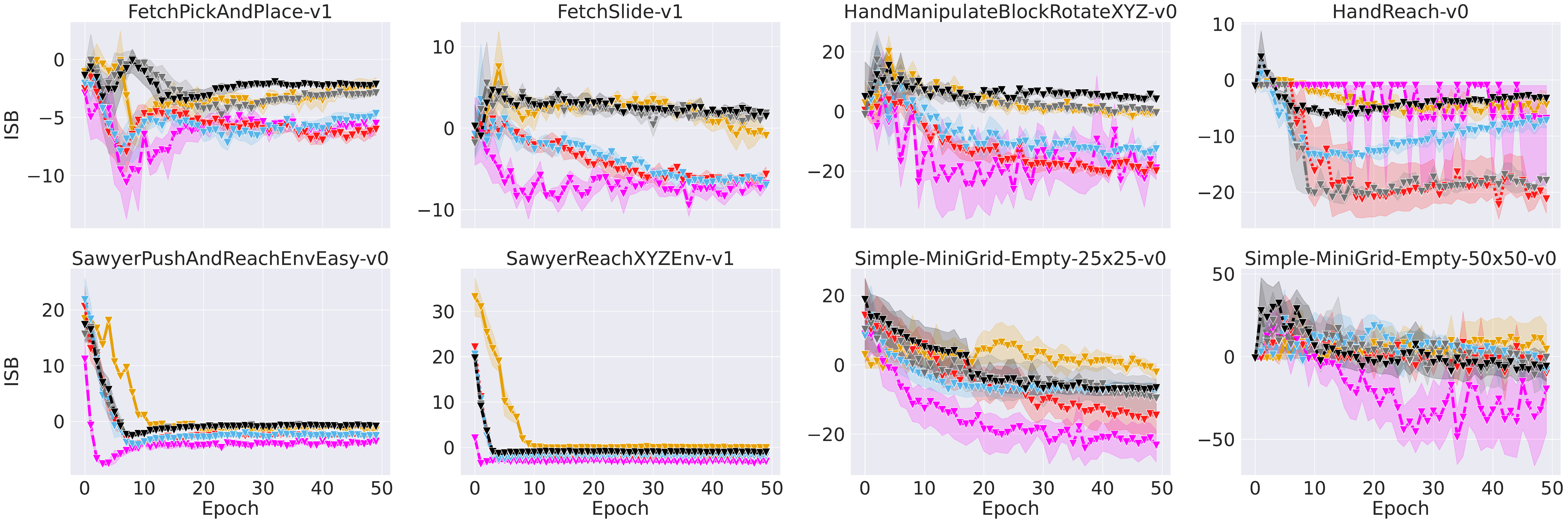}
        \caption{}
        \label{fig:ablation_n7_shooting_bias}
    \end{subfigure}
    \begin{subfigure}[b]{\textwidth}
        \centering
        \includegraphics[width=\textwidth]{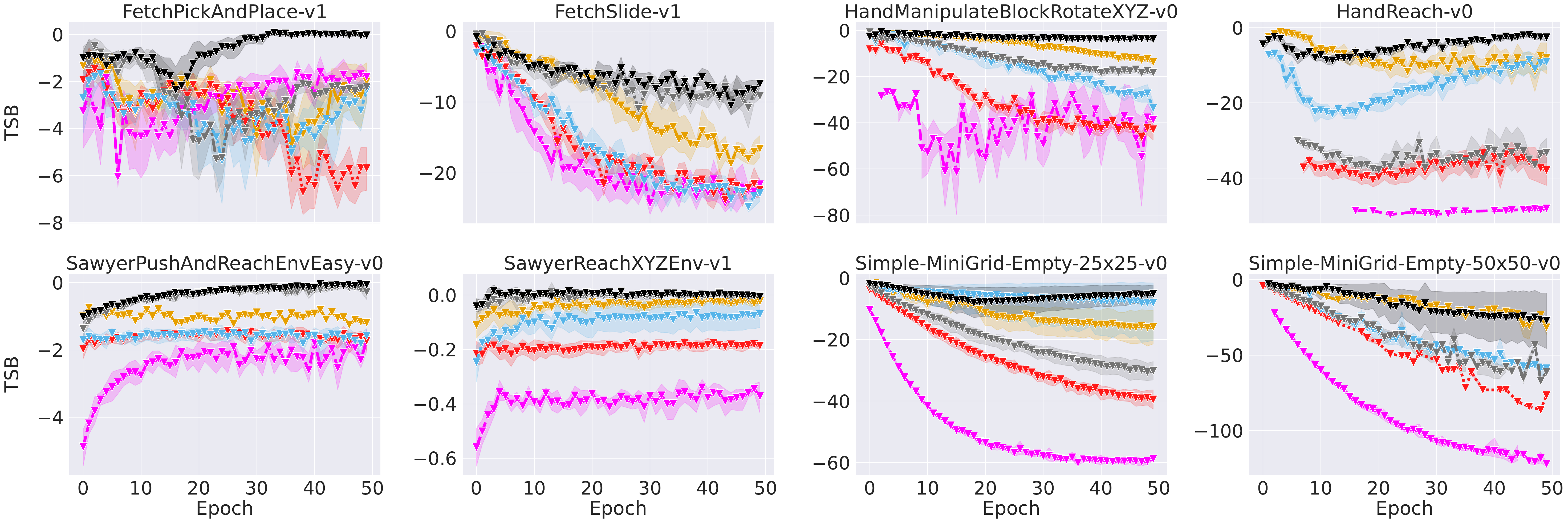}
        \caption{}
        \label{fig:ablation_n7_shifting_bias}
    \end{subfigure}
    \caption{Ablation study of seven-step ($n$=7) GCRL methods on robotic tasks and ten-step (n=10) GCRL methods on grid-world tasks. (a) Success rate. (b) ISB for measuring shooting bias. (c) TSB for measuring shifting bias.}
    \label{fig:ablation_n7}
\end{figure}

Advancing into the ablation study, six methods, namely HER, MHER, MHER($\lambda$), TMHER($\lambda$), QR-MHER, and BR-MHER, are meticulously analyzed. The discerned outcomes are pictorially represented in Fig.~\ref{fig:ablation_n7}.

From Fig.~\ref{fig:ablation_n7}(a), it is evident that MHER($\lambda$) consistently enhances the stability of MHER across diverse tasks and step sizes, thereby providing a more stable performance trajectory. 

Our variant, TMHER($\lambda$), stands as a further consistent enhancement of MHER($\lambda$), universally exhibiting positive advancements in success performance, learning efficiency, and off-policy bias mitigation over MHER($\lambda$) across all cases. This underscores the favorable impact of employing truncated multi-step targets. As delineated in Figs.~\ref{fig:ablation_n7}(a), significant improvements with TMHER($\lambda$) are notably observed in three tasks: \texttt{HandReach-v0} and two grid-world tasks. In these scenarios, actions can inadvertently deflect the agent from goal coordinates, culminating in negative rewards in the bootstrap path and an amplified shifting bias.
In such scenarios, maintaining position at goal states necessitates a self-loop in state transitions, which can exacerbate shifting bias significantly as TD errors increase due to off-policy data. Addressing this issue by switching to single-step learning for transitions among goal states, as facilitated by truncated multi-step targets, results in a marked improvement in task performance.
It is imperative to note that TMHER($\lambda$) outperforms MHER($\lambda$) in \texttt{FetchSlide-v1}, a task where the comprehensive integration of BR-MHER falls short. This observation highlights the potential detrimental impact of quantile regression on the performance within the \texttt{FetchSlide-v1} context. Concurrently, despite the improvements in off-policy biases with TMHER($\lambda$) over MHER($\lambda$), Figs.~\ref{fig:ablation_n7}(b,c) reveal that off-policy biases, as indicated by ISB and TSB, maintain high levels in tasks such as \texttt{FetchSlide-v1} and \texttt{HandManipulateBlockRotateXYZ-v0}. This highlights the circumscribed capacity of TMHER to diminish both shooting and shifting biases.

In the examination of QR-MHER's performance as depicted in Fig.~\ref{fig:ablation_n7}(a), QR-MHER is observed to markedly excel over TMHER($\lambda$) in both \texttt{FetchPickAndPlace-v1} and \texttt{HandManipulateBlockRotateXYZ-v0}. This superiority highlights the advantageous management of shooting bias. Observations from Fig.~\ref{fig:ablation_n7}(b,c) denote that QR-MHER markedly diminishes the off-policy biases of MHER($\lambda$) across all tasks, with the exception of \texttt{HandReach-v0} and grid-world tasks. In these exceptional tasks, TMHER($\lambda$) demonstrates superior bias mitigation capabilities. 
This observed pattern points to a potential limitation in QR-MHER's capability to adequately address the shift in action-values toward biased values within goal states, which can result in unstable learning and reduced efficiency.

Consequently, our BR-MHER amalgamates the advantages of both QR-MHER and TMHER($\lambda$) seamlessly. This integration leads to a superior performance that transcends the peak efficiencies of both QR-MHER and TMHER($\lambda$). In \texttt{HandReach-v0} and grid-world tasks, as depicted in Fig.~\ref{fig:ablation_n7}(a), BR-MHER demonstrates an augmented performance compared to the two variants above. Additionally, the magnitudes of both ISB and TSB are notably reduced, indicative of enhanced off-policy bias mitigation, as evident in Fig.~\ref{fig:ablation_n7}(b,c). Notably, in these tasks where TMHER($\lambda$) surpasses QR-MHER, BR-MHER further diminishes both off-policy biases. This highlights the effective management of both shifting and shooting biases by quantile regression when paired with truncated multi-step targets. Despite a noted performance decline in \texttt{FetchSlide-v1} due to quantile regression, BR-MHER maintains robust performance across a variety of tasks.

To address the fourth research question: A4) Our findings reveal that quantile regression effectively reduces shooting bias and is crucial for enhancing performance in both \texttt{FetchPickAndPlace-v1} and \texttt{HandManipulateBlockRotateXYZ-v0}. Truncated multi-step targets prove to be vital for success in \texttt{HandReach-v0} and grid-world tasks, particularly in scenarios with pronounced shifting bias. Furthermore, the combination of truncated multi-step targets with quantile regression demonstrates an augmented ability to mitigate off-policy bias issues, indicating their synergistic rather than independent functionality.

\section{Conclusion}
This paper provides a robust exploration of the off-policy bias challenges inherent in multi-step GCRL. By uncovering and thoroughly analyzing two types of off-policy biases present in multi-step GCRL, we have shed light on innovative techniques for regulating these biases. This led to the inception of our resilient multi-step GCRL algorithm, BR-MHER. Combined metrics, encapsulating aspects of performance and off-policy bias, robustly showcases the resilience and robustness of our BR-MHER against bias issues, especially as the $n$ increases in the $n$-step targets.
Notably, our experimental evaluations robustly demonstrate the superiority of BR-MHER over the state-of-the-art MHER($\lambda$) and other advanced multi-step baselines for a majority of the GCRL benchmarks in most tasks, both in terms of success rate and learning efficiency. Although our analysis primarily pertains to a deterministic environment, it underscores the potential of harnessing beneficial
off-policy bias instead of avoiding all off-policy bias altogether.

The presented research underpins the critical need for discerning the nature of biases in multi-step GCRL and postulates that not all biases are detrimental. Recognizing and leveraging these biases can be transformative, carving a path for superior, adaptive, and efficient algorithms. It redefines the approach towards bias in multi-step GCRL, potentially unlocking expansive applications across a myriad of complex environments. Moreover, while our approach demonstrates robust performance across a variety of tasks, our observations in certain specific scenarios highlight opportunities for further refinement and enhancement of our strategies. Our ongoing work aims to explore a more robust strategy for off-policy bias mitigation and extend our approach to stochastic environments.

\section*{Appendix}

\appendix
\titleformat{\section}{\large\bfseries}{Appendix \Alph{section}:}{1em}{}
\renewcommand{\theequation}{A\arabic{equation}} 
\renewcommand\thefigure{\Roman{figure}}
\setcounter{equation}{0}
\setcounter{figure}{0}

\section{Background} \label{sect:background}
This appendix offers an in-depth exploration of the technical aspects of TD3, HER, MHER, MHER($\lambda$), MMHER, Retrace($\lambda$), and WGCSL. Additionally, we furnish a detailed derivation of the learning objective of TD error in multi-step GCRL, as delineated in Eq.~(3) from the main manuscript.

\subsection{Twin Delayed Deep Deterministic Policy Gradient (TD3)} \label{sect:td3}
TD3 \citep{fujimoto2018addressing} is an off-policy reinforcement learning method with an actor-critic architecture designed for continuous control tasks. The actor \( \pi_{\psi} \) is parameterized by \( \psi \), and there are two critics \( Q_{\theta_1} \) and \( Q_{\theta_2} \) parameterized by \( \theta_1 \) and \( \theta_2 \), respectively. Given a data tuple \( (s, a, s', r) \), the target value is calculated as
\[ y(s, a, s', r) = r + \gamma \min_{i=1,2} Q_{\bar{\theta}_i}(s', \pi_{\bar{\psi}}(s')) \]
This approach is also known as the Clipped Double Q-learning (CDQ) algorithm used to mitigate over-optimistic bias in the learning process. Unlike traditional actor-critic algorithms that update the actor and critics simultaneously, TD3 proposes to delay the actor updates until the action-value errors are small. This delay is typically implemented by setting the actor update period \( k \) to be larger than one.
The training loss for the critic is represented by the expected \( L2 \) loss between the \( Q \)-value predictions and the target values \( y \) as 
\[ L_{\theta_i} = \mathbb{E}_{(s, a, s', r)\sim\mathcal{B}}[(y(s, a, s', r) - Q_{\theta_i}(s', a))^2],~i\in\{1, 2\} \]
Every \( k \) training steps, the actor is optimized by maximizing the \( Q \)-value of the action output \( \pi_\psi(s) \) for the given states, and the corresponding loss is
\[ L_{\psi} = -\mathbb{E}_{s \sim \mathcal{B}} [Q_{\theta_1}(s, \pi_{\psi}(s))] \]
In the following discussion of algorithms, we omit parameters when they are not necessary, prioritizing clarity and simplicity.

\subsection{Hindsight Experience Replay (HER)}
HER \citep{andrychowicz2017hindsight} proposes to tackle sparse learning signals in sparse-reward settings via relabeling the goal as the actually achieved goals in the future. Consider a goal-conditioned data tuple $(s^\tau_t, a^\tau_t, s^\tau_{t+1}, g^\tau)$, extracted from the trajectory $\tau$. The goals actually achieved in the future constitute the set $\{\phi(s^\tau_{t'})\}_{t'=t+1}^T$. Under HER, the original goal $g^\tau$ is relabeled as $\hat{g}^\tau$, selected from the distribution $p(\{\phi(s^\tau_{t'})\}_{t'=t+1}^T)$. This distribution is dependent on the hindsight relabeling strategy and is typically a uniform distribution over $\{\phi(s^\tau_{t'})\}_{t'=t+1}^T$.

Consequently, the agent learns from the relabeled data tuple $(s^\tau_t, a^\tau_t, s^\tau_{t+1}, \hat{g}^\tau)$ with a recalculated reward $r(s^\tau_{t+1}, \hat{g}^\tau)$. This process ensures that the agent, at some point, receives valuable rewards propagated from the relabeled goal $\hat{g}^\tau$. The target value based on the relabeled goal is computed as:
\begin{equation}
  y(s^\tau_t, a^\tau_t, s^\tau_{t+1}, \hat{g}^\tau) = r(s^\tau_{t+1}, \hat{g}^\tau) + \gamma Q_\pi(s^\tau_{t+1}, \pi(s^\tau_{t+1}, \hat{g}^\tau), \hat{g}^\tau), \label{eq:her}
\end{equation}
where the subscript \( \pi \) of \( Q_\pi \) denotes that the target values for the action-value function are bootstrapped with the action decision from policy \( \pi \). 
With the hindsight relabeling strategy for trajectory $\tau$ at time $t$ represented by a stochastic function $h(t, \tau)$, where the relabeled goal $\hat{g}^\tau$ is sampled from the distribution defined by $h(t, \tau)$, the loss for the goal-conditioned action-value function can be expressed as:

\begin{equation}
    L_{Q_\pi} = \mathbb{E}_{(s^\tau_t, a^\tau_t, s^\tau_{t+1}, g^\tau_t)\sim \mathcal{B}, \hat{g}^\tau\sim h(t, \tau)}[(y(s^\tau_t, a^\tau_t, s^\tau_{t+1}, \hat{g}^\tau) - Q_\pi(s^\tau_t, a^\tau_t, \hat{g}^\tau))^2]. \label{eq:her_critic}
\end{equation}
Then the goal-conditioned policy learns to optimize the expected  $Q$-value function of $\pi(s^\tau_t, \hat{g}^\tau)$ for given states $s^\tau_t$ and corresponding goals $\hat{g}^\tau$ as 
\begin{equation}
        L_{\pi} = \mathbb{E}_{(s^\tau_t, g^\tau_t)\sim \mathcal{B}, \hat{g}^\tau\sim h(t, \tau)}[- Q_\pi(s^\tau_t, \pi(s^\tau_t, \hat{g}^\tau), \hat{g}^\tau_t)]. \label{eq:her_actor}
\end{equation}

\subsection{Multi-Step Hindsight Experience Replay (MHER)}
In comparison to HER, MHER \citep{yang2021bias} integrates HER into multi-step GCRL with multi-step relabeling. Unlike Eq.~(\ref{eq:her}) that calculates the target value on a single-step transition, MHER calculates the target values on $n$-step transitions with the goal $g$ as follows 
\begin{equation}
    y^{(n)}(t, \tau, g) = \sum_{i=0}^{n-1} \gamma^i r(s^\tau_{t+i+1}, g) + \gamma^n Q_\pi(s^\tau_{t+n}, \pi(s^\tau_{t+n}, g), g). \label{eq:mher}
\end{equation}
MHER optimizes the actor and critic in similar ways as Eq.~(\ref{eq:her_critic}) and Eq.~(\ref{eq:her_actor}) except the target values are calculated via Eq.~(\ref{eq:mher}). MHER enables the agent to learn faster from $n$-step targets.

\subsection{MHER(\(\lambda\))}\label{sect:lambda}
Inspired by TD($\lambda$) \citep{sutton2018reinforcement}, MHER($\lambda$) \citep{yang2021bias} provides a novel approach to balance the trade-off between fewer-step targets that come with lower off-policy bias and multi-step targets, which carry more reward information. This balancing act is achieved through the introduction of an exponential decay weight parameter $\lambda$, where $\lambda \in [0, 1]$. The multi-step targets, balanced by the $\lambda$ parameter, are calculated using the following formula:
\begin{equation}
    y^{(n)}_\lambda(t, \tau, g) = \frac{\sum_{i=1}^n\lambda^i y^{(i)}(t, \tau, g)}{\sum_{i=1}^n\lambda^i}. \label{eq:mher_lambda}
\end{equation}
In this formulation, $y^{(n)}_\lambda$ leans towards the one-step target as $\lambda\rightarrow0$, while higher weights are assigned to the $n$-step target as $\lambda$ increases.

\subsection{Model-Based MHER (MMHER)}
MMHER \citep{yang2021bias} calculates on-policy multi-step targets from multi-step transitions generated via a world-model $M$ of the environment. The world-model $M$ is trained by past experiences. Given a one-step transition $(s^\tau_t, a^\tau_t, s^\tau_{t+1})$, the agent simulates $(n-1)$-step on-policy transitions starting from $s^\tau_{t+1}$, following policy $\pi$, in order to pursue the goal $g$ through the world-model $M$. Within these $(n-1)$-step on-policy transitions, the $i$th ($i=1\cdots n-1$) state generated from $M$ is denoted as $M^{\pi_g}_i(s^\tau_{t+1})$, corresponding to the original $s^\tau_{t+1+i}$ in the off-policy data. Then, the multi-step target in MMHER is expressed as
\begin{align}
    y^{(n)}_M(s^\tau_t, a^\tau_t, s^\tau_{t+1}, g) = r(s^\tau_{t+1}, g) +&  \sum_{i=1}^{n-1} \gamma^i r(M^{\pi_g}_i(s^\tau_{t+1}), g)\nonumber\\ 
    +& \gamma^n Q_\pi(M^{\pi_g}_{n-1}(s^\tau_{t+1}), \pi(M^{\pi_g}_{n-1}(s^\tau_{t+1}), g), g).\label{eq:mmher} 
\end{align}
However, $y^{(n)}_M$ actually suffers from model bias caused by the prediction errors of $M$, which can be severe when the problems are complex. To balance the model bias and learning information, \cite{yang2021bias} proposes to integrate $n$-step model-based target and one-step target via a hyper-parameter $\alpha$ as
\begin{equation}
    \bar{y}^{(n)}_M(s^\tau_t, a^\tau_t, s^\tau_{t+1}, g) = \frac{\alpha y^{(n)}_M(s^\tau_t, a^\tau_t, s^\tau_{t+1}, g) + r(s^\tau_{t+1}, g) + \gamma Q_\pi(s^\tau_{t+1}, \pi(s^\tau_{t+1}, g), g)}{1+\alpha}.\label{eq:mix_mmher} 
\end{equation}
The $\bar{y}^{(n)}_M(s^\tau_t, a^\tau_t, s^\tau_{t+1}, g)$ will serve as the final $n$-step targets for the action-value function to learn.

\subsection{Weighted Goal-Conditioned Supervised Learning (WGCSL)}

The Weighted Goal-Conditioned Supervised Learning (WGCSL) algorithm, as delineated by \cite{yang2022rethinking}, stands as a supervised learning methodology designed for the execution of goal-conditioned tasks. This innovative algorithm employs goal relabeling techniques analogous to those used in Hindsight Experience Replay (HER) \citep{andrychowicz2017hindsight}. 

The formal supervised learning objective, \( L_\psi^{WGCSL} \), is expressed as:
\begin{equation}
    L_\psi^{WGCSL} = \mathbb{E}_{(s^\tau_t, a^\tau_t,  g^\tau_t)\sim \mathcal{B}, \phi(s^\tau_i) \sim h(t, \tau)}[w^\tau_{t,i} \log~\pi( a^\tau_t |s^\tau_t, \phi(s^\tau_i))].  \nonumber
\end{equation}

In this equation, \(s^\tau_i\) denotes the prospective state of \(s_t^\tau\) where \(i>t\), with the relabeled goal represented as \(\phi(s^\tau_i)\), and the weight \(w^\tau_{t,i}\) accentuating the prioritization of actions possessing a larger potential. 

The weight \(w^\tau_{t,i}\) is further defined as:
\begin{equation}
w^\tau_{t,i} = \gamma^{i-t} \cdot \text{clip}(\exp(A(s^\tau_t, a^\tau_t, \phi(s^\tau_i))), 0, K) \cdot \epsilon(A(s^\tau_t, a^\tau_t, \phi(s^\tau_i))).\label{eq:wgcsl_weight}
\end{equation} 

Here, \(\gamma^{i-t}\), \(\text{clip}(\exp(A(s^\tau_t, a^\tau_t, \phi(s^\tau_i))), 0, K)\), and \(\epsilon(A(s^\tau_t, a^\tau_t, \phi(s^\tau_i)))\) symbolize the discounted relabeling weight (DRW), goal-conditioned exponential advantage weight (GEAW), and best-advantage weight (BAW) respectively. The DRW prioritizes learning based on the actions of states temporally closer to the relabeled goal. The GEAW emphasizes learning on actions with a higher advantage, employing \(K\) to limit values in extreme cases. The BAW addresses the challenge of multi-modality in goal-conditioned Reinforcement Learning (RL), guiding the policy towards the modal with the maximal return, circumventing convergence to a weighted average of multiple modals. BAW is specifically formulated as:
\begin{equation}
    \epsilon(A(s^\tau_t, a^\tau_t, \phi(s^\tau_i))) = 
    \begin{cases} 
        1, & \text{if } A(s^\tau_t, a^\tau_t, \phi(s^\tau_i)) > \hat{A} \\
        \epsilon_{\text{min}}, & \text{otherwise}
    \end{cases} \nonumber
\end{equation}
Within this context, \(\hat{A}\) denotes a threshold calculated from recent advantage statistics, while \(\epsilon_{\text{min}}\) represents a minimal positive value.

\subsection{Retrace($\lambda$)}
Retrace($\lambda$) is a reinforcement learning algorithm specifically designed for off-policy correction of the value estimates \citep{munos2016safe}. The algorithm employs importance sampling (IS) for correction, incorporating a safety mechanism to ensure the stability and robustness of learning. This mechanism involves truncating the importance sampling ratios to mitigate excessive variance in the estimates.

The $n$-step goal-conditioned target value based on Retrace($\lambda$) is calculated as
\begin{equation}
    y^{(n)}_{\text{ret}}(t, \tau, g) = \sum_{i=1}^{n} \gamma^{i-1} \left(\prod_{s=1}^{i-1} c_s\right)\left(r(s^\tau_{t+i}, g) + \gamma \mathbb{E}_{\pi} [Q_\pi(s^\tau_{t+i}, \cdot)] - \gamma c_{i}  Q(s^\tau_{t+i}, a^\tau_{t+i})\right), \nonumber
\end{equation}
where $c_i$ is the truncated importance sampling ratio. It is defined that $\prod_{s=1}^0 c_s=1$ and $c_{n}=0$ as the experience beyond the $n$ steps is truncated. Specifically, the importance sampling ratio $c_t$ is:
\begin{equation}
    c_i = \lambda \min \left( 1, \frac{\pi(a^\tau_{t+i}|s^\tau_{t+i})}{\mu(a^\tau_{t+i}|s^\tau_{t+i})} \right), \label{eq:import_ratio}
\end{equation}
where $\mu(a_t|s_t)$ is the behavior policy that generated the action $a_t$ at state $s_t$. The truncated importance sampling enables the agent to safely use off-policy data to correct the estimates.
\subsection{TD Error Learning Objective in Multi-Step GCRL}
This section meticulously delineates the derivation of the objective, depicted as Eq.~(3) in the main text.

\begin{align}
&Q_\theta(s^\tau_{t}, a^\tau_{t}, g) \nonumber\\
=& ~r_t + \gamma Q_\btheta(s^\tau_{t+1}, \pi_\bpsi(s^\tau_{t+1}, g), g) - \delta_\theta(s^\tau_{t}, a^\tau_{t}, g) \nonumber\\
=& ~r_t + \highlight{\gamma Q_\btheta(s^\tau_{t+1}, a^\tau_{t+1}, g)}  - \gamma A_\btheta(s^\tau_{t+1}, a^\tau_{t+1}, g) - \delta_\theta(s, a, g)\label{eq:ill_posed_bias_1} \\
=& ~r_t + \highlight{\gamma \left [ r_{t+1} + \gamma Q_\btheta(s^\tau_{t+2},\pi_\bpsi(s^\tau_{t+2}, g), g) - \delta_\btheta(s_{t+1}^\tau, a_{t+1}^\tau, g)\right ]} - \gamma A_\btheta(s^\tau_{t+1}, a^\tau_{t+1}, g) \nonumber\\
\phantom{=}&  - \delta_\theta(s^\tau_{t}, a^\tau_{t}, g) \label{eq:ill_posed_bias_2} \\
=& ~\cdots \nonumber \\
=& ~\sum_{i=0}^{n-1}\gamma^i r_{t+i} + \gamma^n Q_\btheta(s^\tau_{t+n}, \pi_\bpsi(s^\tau_{t+n}, g), g) - \sum_{i=1}^{n-1} \gamma^i [A_\btheta(s^\tau_{t+i}, a^\tau_{t+i}, g) 
+ \delta_\btheta(s^\tau_{t+i}, a^\tau_{t+i}, g)]  \nonumber\\
\phantom{=}&  -  \delta_\theta(s^\tau_{t}, a^\tau_{t}, g) \nonumber\\
=& ~y^{(n)}(t, \tau, g)   - \sum_{i=1}^{n-1} \gamma^i \left [A_\btheta(s^\tau_{t+i}, a^\tau_{t+i}, g) + \delta_\btheta(s^\tau_{t+i}, a^\tau_{t+i}, g) \right] -  \delta_\theta(s^\tau_{t}, a^\tau_{t}, g). \label{eq:td_objective}
\end{align}

In Eq.~(\ref{eq:ill_posed_bias_1}), the highlighted term, $\gamma Q_\btheta(s^\tau_{t+1}, a^\tau_{t+1}, g)$, is purposefully added and then subtracted by $Q_\btheta(s^\tau_{t+1}, \pi_\bpsi(s^\tau_{t+1}, g), g)$, resulting in $\gamma A_\btheta(s^\tau_{t+1}, a^\tau_{t+1}, g)$, thereby initiating the unrolling process. Following this, Eq.~(\ref{eq:ill_posed_bias_2}) showcases the expanded form of the previous term, elucidating the unwinding of the action-value function's recursive definition. This expansion includes the subsequent reward $r_{t+1}$, the ensuing action-value function $Q_\btheta(s^\tau_{t+2},\pi_\bpsi(s^\tau_{t+2}, g), g)$, and the TD error $\delta_\btheta(s_{t+1}^\tau, a_{t+1}^\tau, g)$. Consequently, Eq.(3) from the main text is directly extrapolated from Eq.~(\ref{eq:td_objective}).

\section{Experimental and Implementation Details}\label{sect:exp_imp_details}
In this appendix, we furnish extensive details regarding our experimental tasks, implementation specifics, hardware requisites of our experiments, and information pertinent to our released code.
\subsection{Environments and Tasks}
In this section, we provide comprehensive elaboration on the four unique task types incorporated into our experimental procedures.
\paragraph{Fetch Tasks} The Fetch robotic arm \cite{plappert2018multi} represents an advanced, seven-degree-of-freedom (7-DoF) robotic arm outfitted with a two-pronged parallel gripper. The goals are denoted by a 3-dimensional vector that specifies the Cartesian coordinates of the target endpoint. A goal is deemed successfully achieved if the achieved goal resides within a 5cm radius of the intended coordinates. Notably, the arm features a four-dimensional continuous action, which encompasses three Cartesian dimensions to regulate the end effector's movement, along with an additional dimension dedicated to controlling the gripper. For the experimental tasks utilized for our study: i) \texttt{FetchPickAndPlace}: The agent is required to grasp the target object,  relocate it to the goal position and sustain its position until the termination of the task; ii) \texttt{FetchSlide}: The task requires the agent to slide a small puck across the table to a goal position that is outside of the reach of the robot, and ensure the object remains within the goal location at the end of the task.

\paragraph{Hand Tasks} The anthropomorphic robotic \cite{plappert2018multi} hand is a 24-DoF robot hand with a 20-dimensional action space. For the tasks used in our experiments: i) \texttt{HandReach}: The goal of the task is a 15-dimensional vector corresponding to the desired Cartesian positions of all five fingertips. The agent is required to manipulate the robot to reach the states where the mean distance between five fingertips and the goal positions is less than 1cm. Its state is a 63-dimensional vector; ii) \texttt{HandManipulateBlockRotateXYZ}: The agent is required to manipulate a block at the palm of the robotic hand to achieve a target pose. The goal is a 7-dimensional vector including 3-dimensional target positions and 4-dimensional target rotations. The agent succeeds in this task if the block differs from the target rotation within 0.1 rad. Its state space is 61-dimensional.

\paragraph{Sawyer Tasks} The Sawyer robotic arm \cite{nair2018visual} is a 7-DoF robotic arm. Its end-effector (EE) is constrained to a 2-dimensional rectangle parallel to a table. The movements and positioning of this EE are dictated by actions relayed through a motion capture (mocap) system.  For the tasks used in our experimental study: i) \texttt{SawyerPushAndReachEnvEasy}: This task necessitates the agent to push a small puck on the table to a target XY position specified by the 2-dimensional goal. The task operates within a 2-dimensional action space corresponding to the X and Y axes.  The state space for this task, encapsulated in five dimensions, encompasses both the 3-dimensional coordinates of the EE and the 2-dimensional positioning of the puck; ii) \texttt{SawyerReachXYZEnv}: This task demands precise manipulation of the robotic arm's EE towards a designated position indicated by Cartesian coordinates. Both the state and the goal spaces are 3-dimensional, encompassing the current and target coordinates of the EE. The action associated with this task is a 3-dimensional continuous variable, each representing the velocity along an axis.

\paragraph{Grid-World Task} The grid-world tasks employed in this study have been adapted from \cite{gym_minigrid} to suit GCRL contexts. In these tasks, the current state is defined by the agent's current 2-dimensional coordinates, while the goal corresponds to the desired coordinates the agent aims to reach. For these grid-world tasks, we set the agent's action space to consist of five discrete actions: upward movement, downward movement, leftward movement, rightward movement, and maintaining the current position. The tasks \texttt{Simple-MiniGrid-Empty-25$\times$25} and \texttt{Simple-MiniGrid-Empty-50$\times$50} are differentiated solely by scale, containing 25 and 50 grids along each edge respectively.

The horizon $T$ varies based on the specific task at hand. For instance, the horizon for tasks such as \texttt{FetchPickAndPlace}, \texttt{FetchSlide}, and \texttt{HandReach} is set at 50. In contrast, the tasks \texttt{HandManipulateBlockRotateXYZ}, \texttt{SawyerPushAndReachEnvEasy}, and \texttt{SawyerReachXYZEnv} have a larger horizon of 100. In the context of grid-world tasks, the horizon is determined as three times the scale of their respective scales. Consequently, this results in a horizon of 75 for the task \texttt{Simple-MiniGrid-Empty-25$\times$25}, and extends to 150 for the task \texttt{Simple-MiniGrid-Empty-50$\times$50}.

\subsection{Implementation Details}
In this section, we detail the implementation of our TD3, GCRL, and WGCSL methods, outlining the network configurations and hyperparameters used.

\paragraph{TD3} TD3 serves as the primary framework for each GCRL method in our experiments, utilizing CDQ, target networks and delayed actor updates as decribed in Sect.~\ref{sect:td3} to ensure the stability of the learning process. The hyper-parameters employed in the experiments are shown in Table~\ref{tb:network_config}. Notably, in experiments on grid-world tasks, which operate within a discrete action space, we employ the GumbelSoftMax activation \citep{jang2016categorical} to the policy network outputs. This approach ensures the action-value function's differentiability relative to the actor outputs within these contexts.  Observations and goals, serving as inputs to the actor and critic, are normalized based on historical statistics. Throughout the training phase, the target networks undergo soft updates with the parameters of the present actor and critic networks. Each epoch contains varying numbers of cycles depending on the task: Fetch and Hand tasks comprise 50 cycles per epoch, while Sawyer and grid-world tasks include 10. For exploration, we apply $\epsilon$-greedy strategies and additionally integrate Gaussian action noises into the training rollouts of robotic tasks.

\paragraph{GCRL Methods} The primary distinctions among GCRL methods stem from the calculation of target values for the critic, as depicted in Section~\ref{sect:background}. All GCRL methods share the same relabeling probability of $p_h=0.8$, implying an average relabeling of four out of five data entries.  For MMHER, we set $\alpha=0.5$ in Eq.~(\ref{eq:mmher}) and utilize the configurations are outlined in Table~\ref{tb:mmher_world_model}, following the suggestions from \cite{yang2021bias}. For both IS-MHER based on Retrace($\lambda$) and MHER($\lambda$), we assign a value of 0.7 to $\lambda$ in both Eq.~(\ref{eq:import_ratio}) and Eq.(\ref{eq:mher_lambda}). In IS-MHER where the importance sampling necessiates a stochastic policy, we establish the stochstic policy via adding a clipped Gaussian distibuted noises $\epsilon_a \sim clip(\mathcal{N}(0, 0.2), -0.5, 0.5)$ to the actions.  Each experiment is conducted with five distinct random seeds—$\{111, 222, 333, 444, 555\}$—and the statistics of the results are reported.

\paragraph{WGCSL}
Given the focus of WGCSL on supervised learning, we opt not to integrate TD3 into its framework. Our experimental setup is based on the source code provided by the author \citep{yang2022rethinking}. The $K$ used for the weight DRW in Eq.~(\ref{eq:wgcsl_weight}) is set to 10. Apart from the non-application of TD3-specific parameters and the substitution of Softmax for GumbelSoftmax in grid-world tasks, all other hyperparameters are consistent with those utilized in other GCRL methods, see Table~\ref{tb:network_config}.

In general, we primarily adhere to the hyper-parameters proposed in the codebase \citep{yang2021bias,yang2022rethinking} to maintain consistency and avoid parameter tuning. This decision stems from our central aim: to determine if our methodologies can surpass both baseline and contemporary state-of-the-art methods when subject to identical conditions. Importantly, our strategies are tailored to tackle the root causes of bias issues, with the intention to elevate performance through an intrinsic comprehension of these challenges.

\begin{table}
  \vspace{11pt} 
  \caption{The hyper-parameters used for TD3}
  \vspace{11pt} 
  \label{tb:network_config}
  \centering
  \begin{tabular}{ll}
    \toprule
    Hyper-parameter  &   Value \\
    \midrule
    Actor learning rate & 0.001 \\
    Critic learning rate & 0.001  \\
    Optimizer & Adam \citep{kingma2014adam} \\
    Buffer size & $10^6$  \\
    Quadratic penalty on actions & 1.0 \\
    Observation clip range & [-200, 200]  \\
    Normalized observation clip range & [-5, 5]  \\
    Random init episodes & 100 (500 for MMHER) \\
    Batch size & 1024 \\
    Test episodes per epoch & 120 \\
    Discount factor $\gamma$ & $1/T$  \\
    $\epsilon$ for $\epsilon$-greedy exploration & 0.3\\
    Gaussian action noise std & 0.2 (robotics) \\
    Relabeling probability & 0.8 \\
    Episodes per cycle & 12 \\
    Training batches per cycle & 40  \\ 
    Target network update proportion & 0.005 \\
    Delayed actor update interval $k$ & 2 \\
    Number of hidden layers & 3   \\
    Hidden layer size & 256 (robotics), 512 (grid)   \\
    Hidden layer activation     & ReLU   \\
    Actor output activation     & Tanh (robotics) \\ 
    &Softmax + GumbelSoftmax (grid-world) \\
    Training cycles per epoch & 50 (Hand and Fetch tasks)\\
    & 10 (Sawyer and grid-world tasks) \\
    \bottomrule
  \end{tabular}
\end{table}

\begin{table}
  \vspace{11pt}
  \caption{The hyper-parameters of the world-model used for MMHER}
  \vspace{11pt}
  \label{tb:mmher_world_model}
  \centering
  \begin{tabular}{ll}
    \toprule
    Hyper-parameter  &   Value \\
    \midrule
    Learning rate  & 0.001 \\
    Optimizer & Adam \\
    Number of hidden layers & 8 \\
    Hidden layer size & 256 \\
    Batch size & 512 \\
    Warmup training times & 500  \\
    \bottomrule
  \end{tabular}
\end{table}

\subsection{Hardware}
In conducting each experiment, we utilized the processing power of a V100 GPU coupled with eight CPU cores, facilitating efficient model training and comprehensive data collection from environments.
\subsection{Code}
We have adopted a modified OpenAI Baselines codebase \citep{baselines}, adhering to the guidelines put forth by \cite{yang2021bias,yang2022rethinking}. Tensorflow \citep{abadi2016tensorflow} serves as our chosen deep learning framework for training. Recognizing the pivotal role of reproducibility in scientific endeavors, we have rendered our code publicly accessible. Detailed instructions to aid in the effective usage of the code are provided. This information can be accessed at \url{https://github.com/BR-MHER/BR-MHER}.

\begin{figure}[ht]
    \centering
    \begin{subfigure}[b]{\textwidth}
        \centering
        \includegraphics[width=\textwidth]{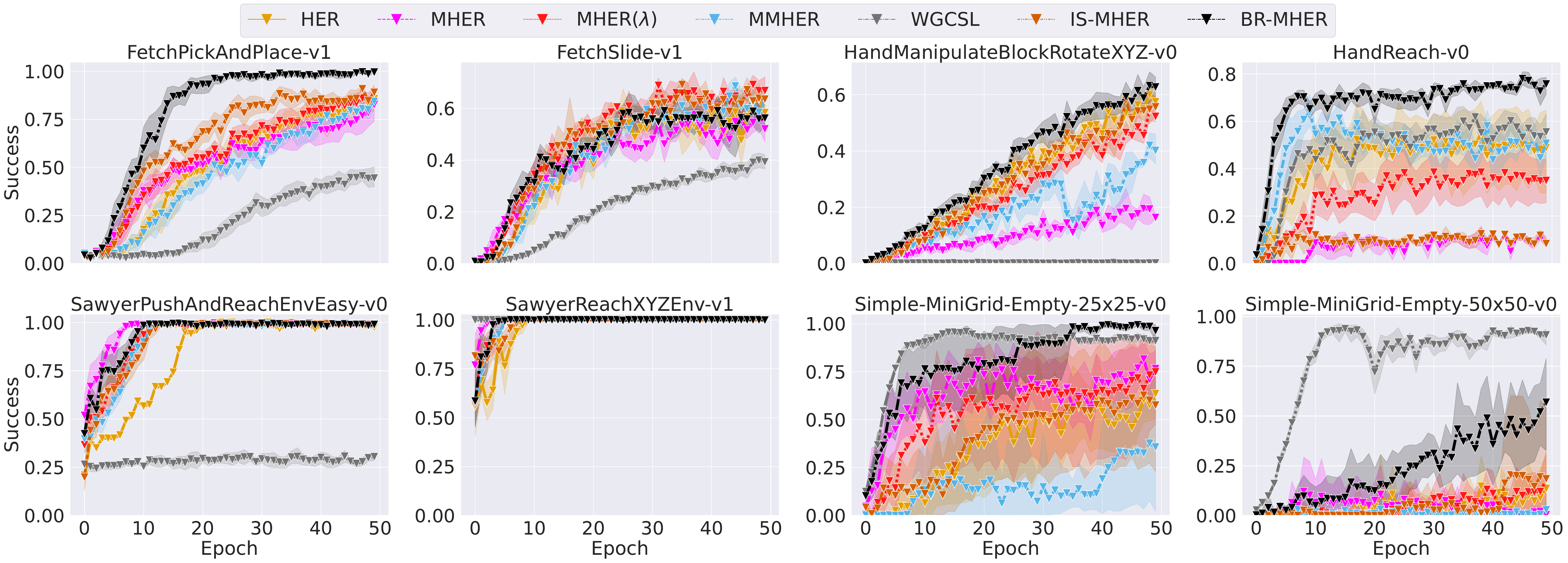}
        \caption{}
        \label{fig:br_mher_n3_success}
    \end{subfigure}
    \begin{subfigure}[b]{\textwidth}
        \centering
        \includegraphics[width=\textwidth]{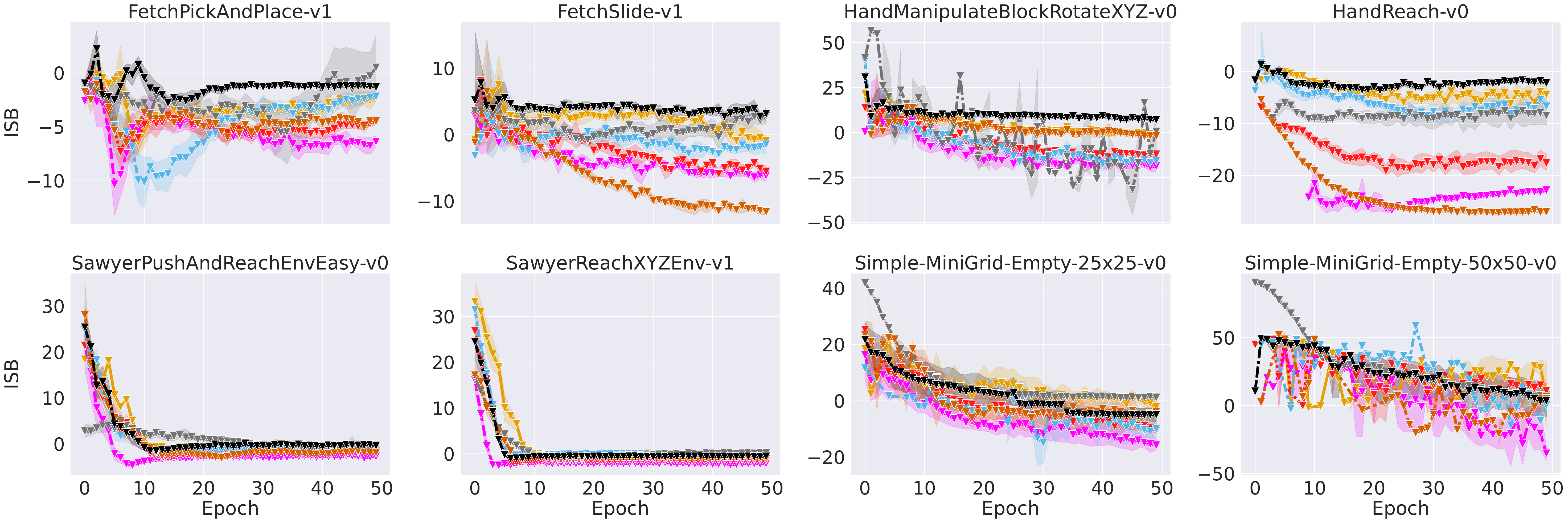}
        \caption{}
        \label{fig:br_mher_n3_shooting_bias}
    \end{subfigure}
    \begin{subfigure}[b]{\textwidth}
        \centering
        \includegraphics[width=\textwidth]{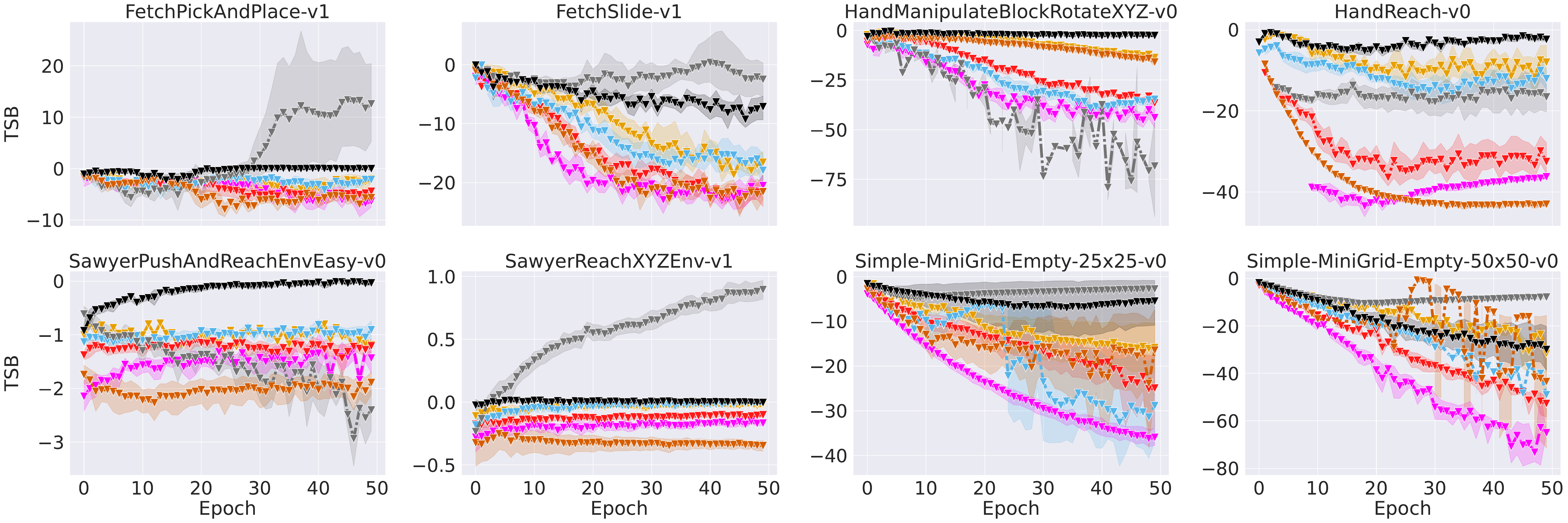}
        \caption{}
        \label{fig:br_mher_n3_shifting_bias}
    \end{subfigure}
    \caption{Comparative study of three-step ($n$=3) GCRL methods on both robotic tasks and grid-world tasks. (a) Success rate. (b) ISB for measuring shooting bias. (c) TSB for measuring shifting bias.}
    \label{fig:br_mher_n3}
\end{figure}

\begin{figure}[t]
    \centering
    \begin{subfigure}[b]{\textwidth}
        \centering
        \includegraphics[width=\textwidth]{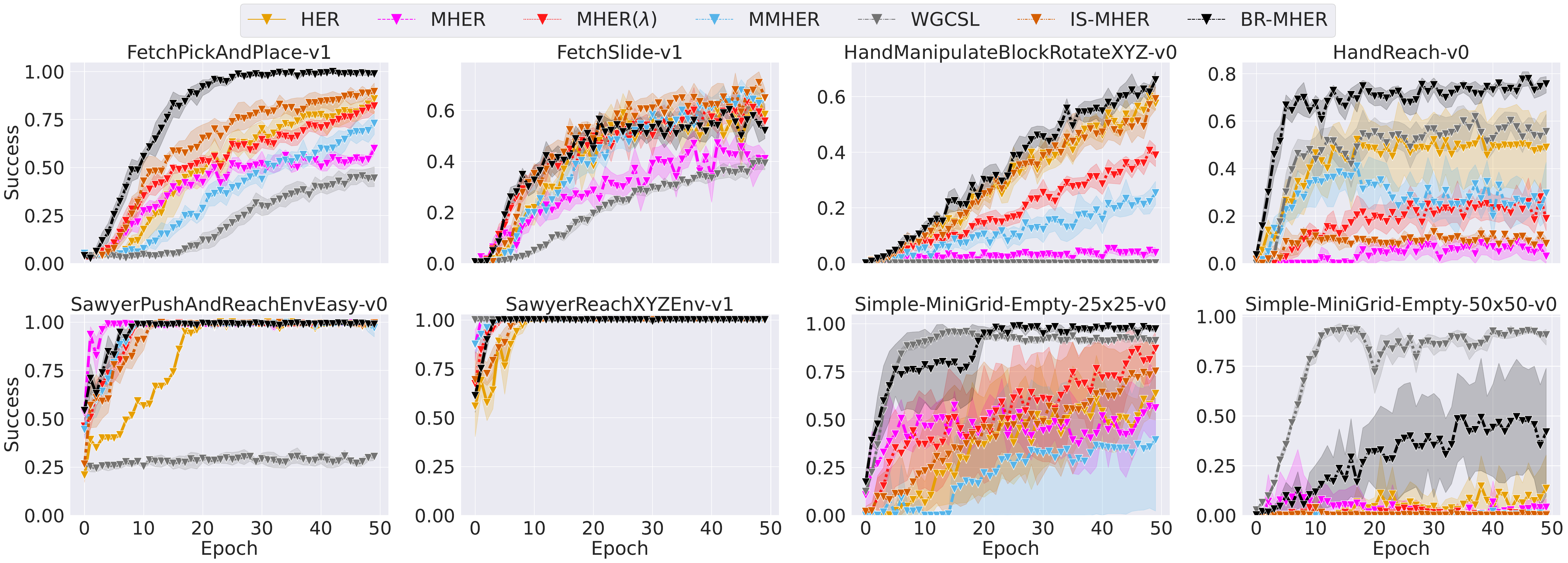}
        \caption{}
        \label{fig:br_mher_n5_success}
    \end{subfigure}
    \begin{subfigure}[b]{\textwidth}
        \centering
        \includegraphics[width=\textwidth]{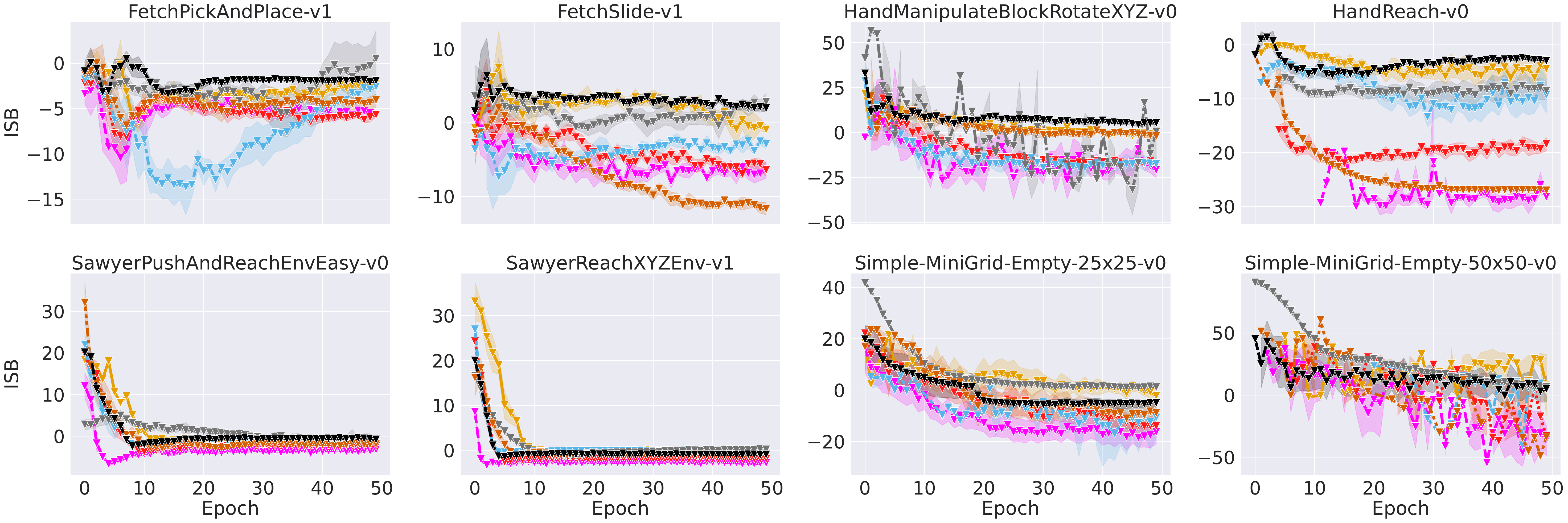}
        \caption{}
        \label{fig:br_mher_n5_shooting_bias}
    \end{subfigure}
    \begin{subfigure}[b]{\textwidth}
        \centering
        \includegraphics[width=\textwidth]{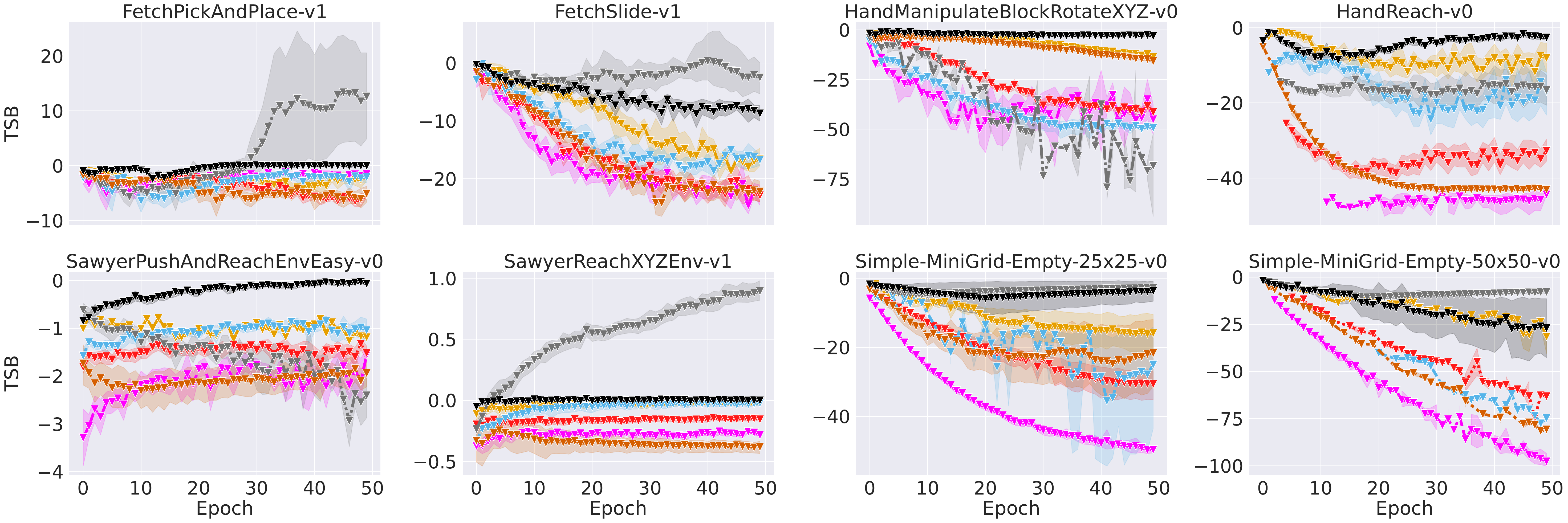}
        \caption{}
        \label{fig:br_mher_n5_shifting_bias}
    \end{subfigure}
    \caption{Comparative study of five-step ($n$=5) GCRL methods on both robotic tasks and grid-world tasks. (a) Success rate. (b) ISB for measuring shooting bias. (c) TSB for measuring shifting bias.}
    \label{fig:br_mher_n5}
\end{figure}

\section{Extended Experiments}\label{sect:ext_exp}
In this appendix, we first enrich the comparative analysis of the BR-MHER application, initially presented in the main text, by conducting a deeper examination of both the three-step ($n$=3) and five-step ($n$=5) GCRL methods applied to the same set of tasks. Following this, we conduct an ablation study to elucidate the components within BR-MHER.
\subsection{More Results of BR-MHER}

This section delineates the remaining comparative analysis of BR-MHER for $n$=3 and $n$=5, as illustrated in Figs.~\ref{fig:br_mher_n3} and \ref{fig:br_mher_n5}.

The robust advantages delineated in the main text persist for smaller step sizes such as 3 or 5. It is observed that BR-MHER consistently mitigates off-policy bias, as evidenced by the ISB and TSB levels in Figs.~\ref{fig:br_mher_n3}(b, c) and \ref{fig:br_mher_n5}(b,c). Here, we can investigate a new research question following the three in the main text: \textbf{Q5)} How does the performance of BR-MHER change as the step-size of multi-step GCRL changes? How is it compared to other GCRL methods?  Instead of performing the one-to-one comparison as the main text, we answer in a general perspective.

As depicted in Figs.~\ref{fig:br_mher_n3}(a) and \ref{fig:br_mher_n5}(a), BR-MHER overtakes the learning efficiency of HER in \texttt{HandManipulateBlockRotateXYZ-v0}, although it aligns with HER's efficiency in a seven-step learning context. This trend highlights that BR-MHER is similarly impacted by the detrimental effects of off-policy biases with increasing step-size. Concurrently, BR-MHER reaps greater advantages than other GCRL methods in grid-world tasks with step-size augmentation. No significant performance shifts are noted for other tasks. An analysis of the performance differential among various multi-step GCRL methods reveals expanding gaps in \texttt{FetchPickAndPlace-v1}, Hand tasks, and grid-world tasks. This observation corroborates the enhanced robustness and resilience of BR-MHER to alterations in step-size and escalating off-policy biases.

Moreover, we find the performance does not decrease monotonically as the step-size increases. Each multi-step GCRL method seems to have the best step-size for learning. Take the \texttt{Simple-MiniGrid-Empty-25x25-v0} for example,  MHER($\lambda$) reaches around 75\% success rates with three-step learning, arising to 85\% with five-step while dropping to 40\% with ten-step learning. In contrast, BR-MHER can stably reach 100\%. In this case, BR-MHER offers an robust option to the parameter of step-size. We also notice that there is a special case in \texttt{HandManipulateBlockRotateXYZ-v0} that IS-MHER keeps the same performance as HER when $n=$3, 5 and 7. It probably reduce the multi-step learning to single-step learning via importance sampling due to the relatively low entropy of the policy.

Furthermore, the analysis discerns a non-monotonic performance degradation with step-size amplification. Each multi-step GCRL method appears to have an optimal step-size for learning. For instance, in the \texttt{Simple-MiniGrid-Empty-25x25-v0} task, MHER\((\lambda)\) records approximately 75\% success with three-step learning, which escalates to 85\% with five-step learning, but falls to 40\% with ten-step learning. Contrarily, BR-MHER consistently achieves a 100\% success rate, underscoring its robustness relative to step-size parameterization. An anomalous observation in \texttt{HandManipulateBlockRotateXYZ-v0} shows that IS-MHER mirrors HER's performance for \(n=\)3, 5, and 7, potentially signifying a transition from multi-step to single-step learning via importance sampling, attributed to the relatively low entropy of the policy.

\subsection{Ablation Study}
\begin{figure}[t]
    \centering
    \begin{subfigure}[b]{\textwidth}
        \centering
        \includegraphics[width=\textwidth]{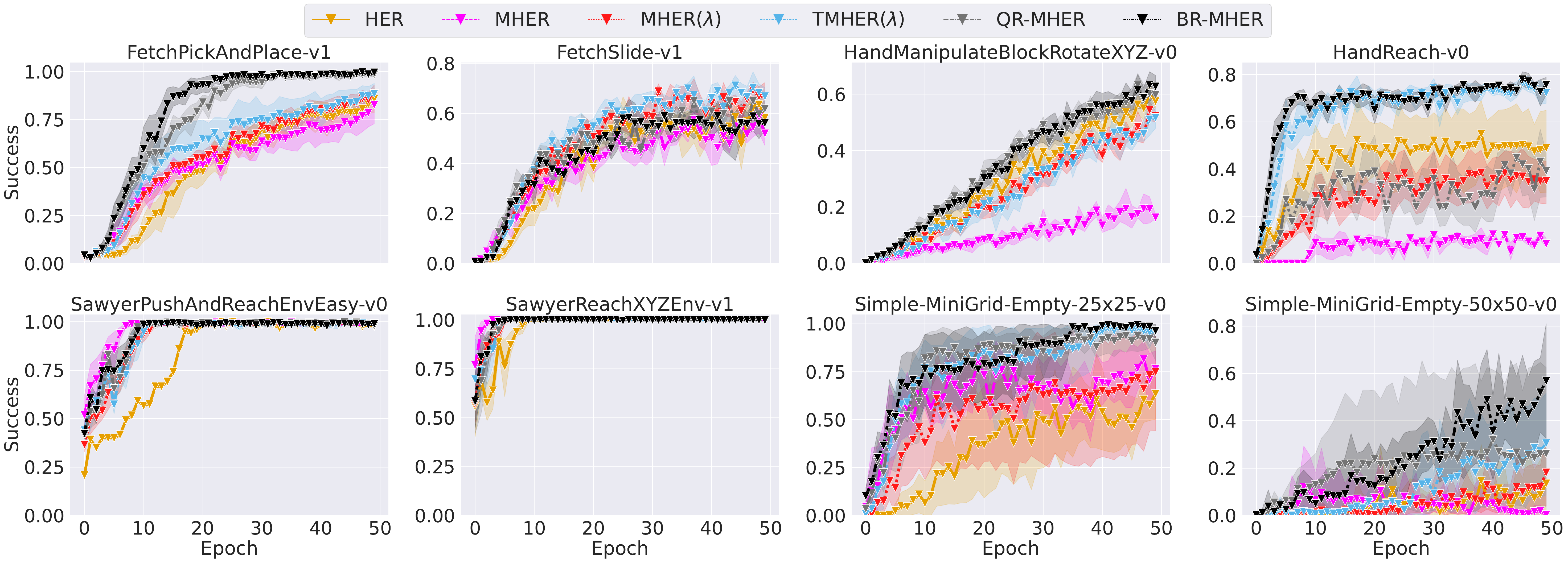}
        \caption{}
        \label{fig:ablation_n3_success}
    \end{subfigure}
    \begin{subfigure}[b]{\textwidth}
        \centering
        \includegraphics[width=\textwidth]{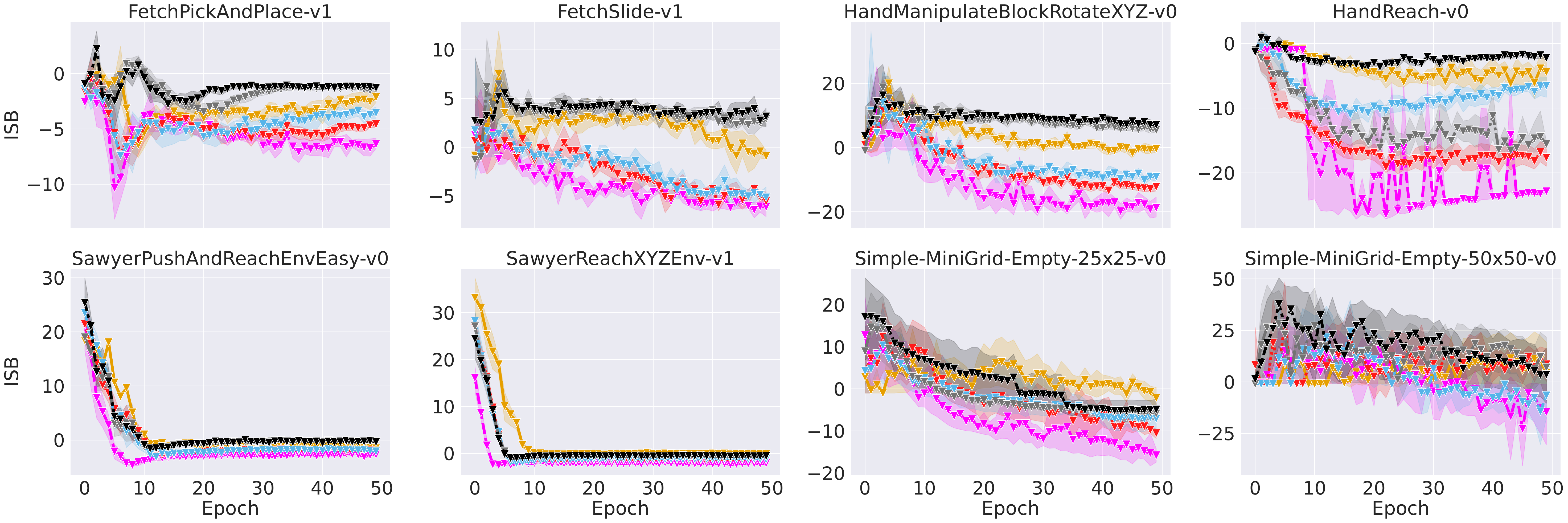}
        \caption{}
        \label{fig:ablation_n3_shooting_bias}
    \end{subfigure}
    \begin{subfigure}[b]{\textwidth}
        \centering
        \includegraphics[width=\textwidth]{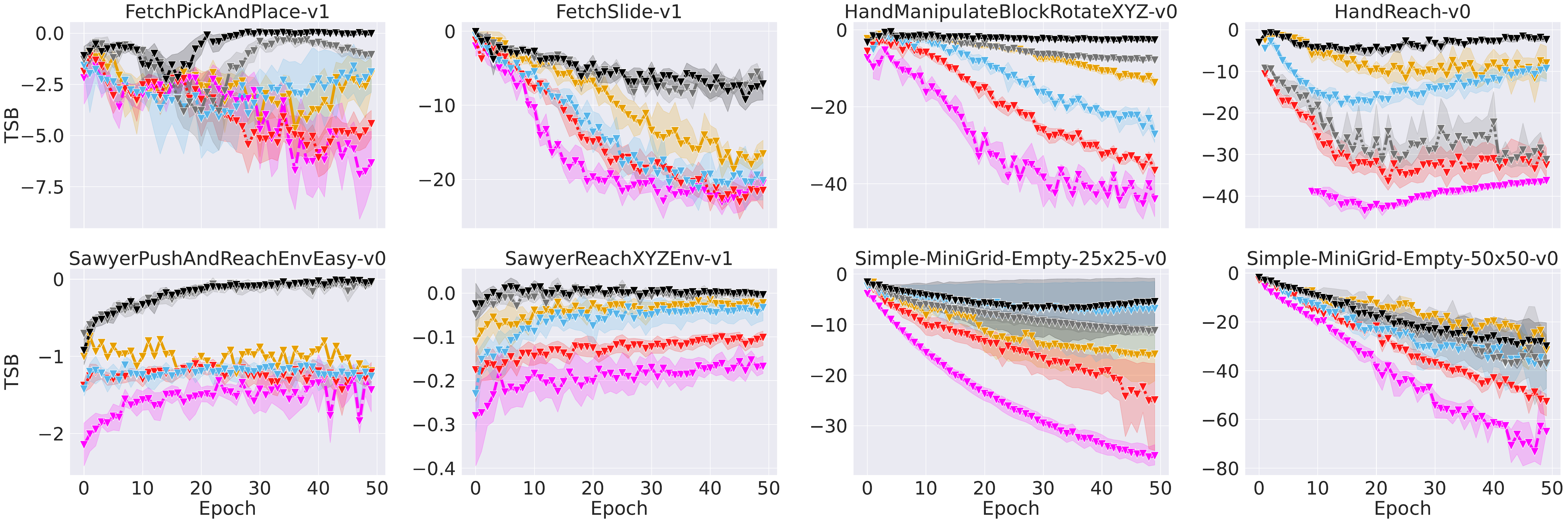}
        \caption{}
        \label{fig:ablation_n3_shifting_bias}
    \end{subfigure}
    \caption{Ablation study of three-step ($n$=3) GCRL methods on both robotic tasks and grid-world tasks. (a) Success rate. (b) ISB for measuring shooting bias. (c) TSB for measuring shifting bias.}
    \label{fig:ablation_n3}
\end{figure}

\begin{figure}[t]
    \centering
    \begin{subfigure}[b]{\textwidth}
        \centering
        \includegraphics[width=\textwidth]{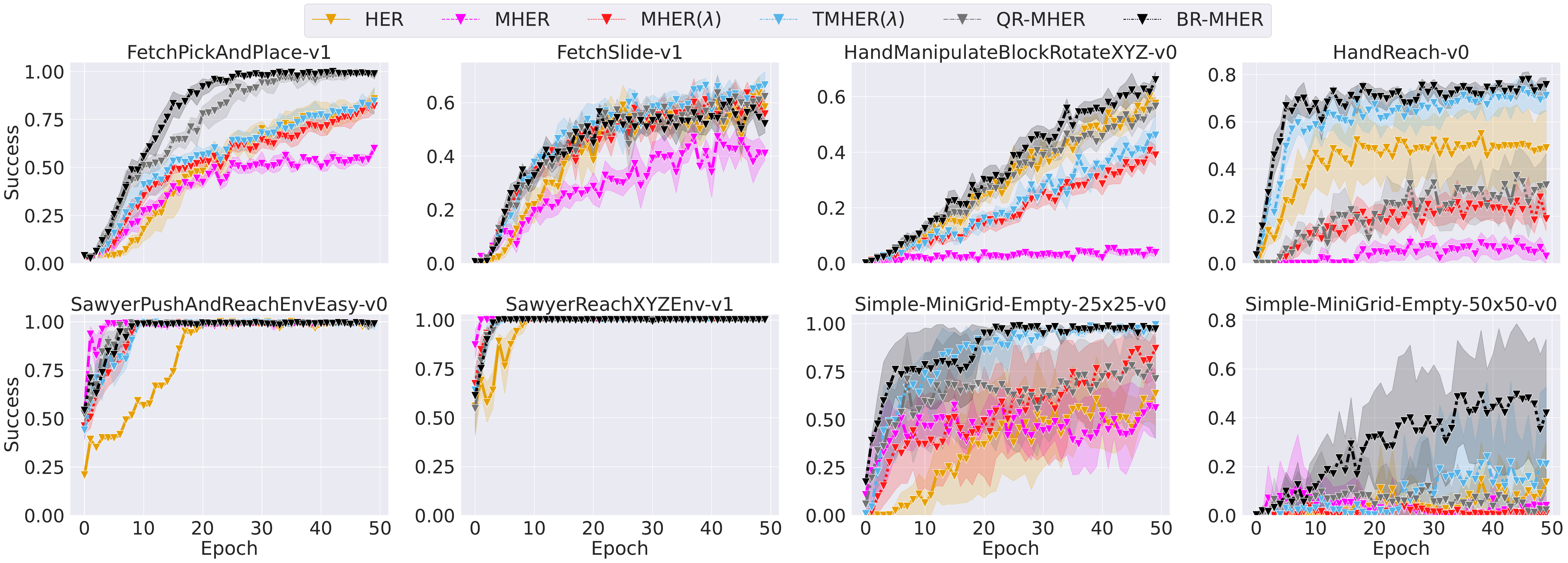}
        \caption{}
        \label{fig:ablation_n5_success}
    \end{subfigure}
    \begin{subfigure}[b]{\textwidth}
        \centering
        \includegraphics[width=\textwidth]{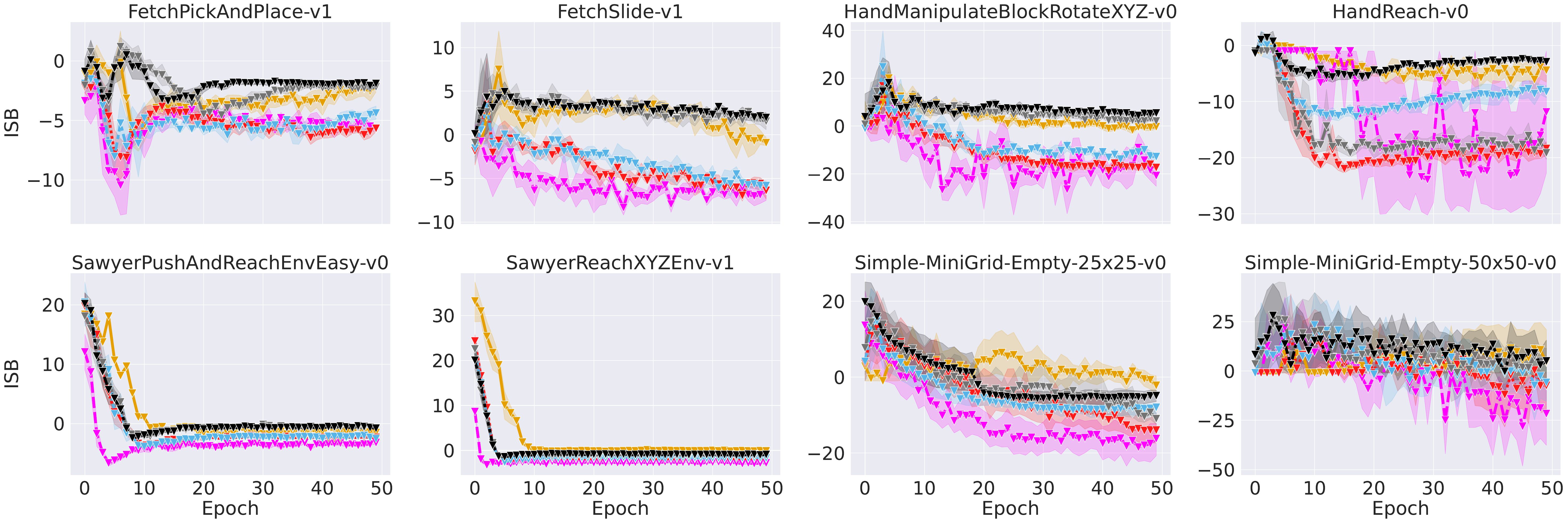}
        \caption{}
        \label{fig:ablation_n5_shooting_bias}
    \end{subfigure}
    \begin{subfigure}[b]{\textwidth}
        \centering
        \includegraphics[width=\textwidth]{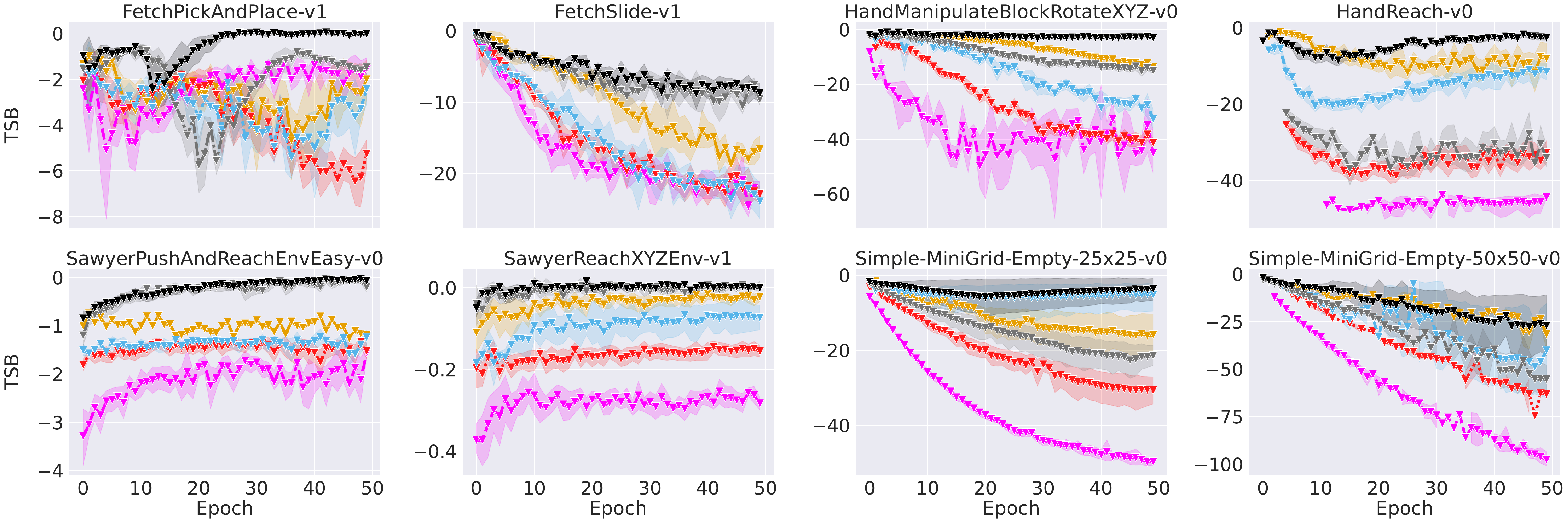}
        \caption{}
        \label{fig:ablation_n5_shifting_bias}
    \end{subfigure}
    \caption{Ablation study of five-step ($n$=5) GCRL methods on both robotic tasks and grid-world tasks. (a) Success rate. (b) ISB for measuring shooting bias. (c) TSB for measuring shifting bias.}
    \label{fig:ablation_n5}
\end{figure}

In this section, we extend the ablation study analysis of our methods for \( n = 3 \) and \( n = 5 \), as showcased in Figs.~\ref{fig:ablation_n3} and \ref{fig:ablation_n5}. The findings discussed in the main text are largely consistent with these smaller step sizes.

However, a notable exception is observed in Fig.~\ref{fig:ablation_n3}(a), where QR-MHER shows enhanced learning efficiency compared to TMHER(\(\lambda\)) in grid-world tasks with \( n = 3 \). Yet, as \( n \) increases, this efficiency advantage diminishes, with TMHER(\(\lambda\)) outperforming QR-MHER. This shift highlights the increasing severity of shifting bias with larger step sizes and underscores the limitations of QR-MHER in addressing this bias alone. It emphasizes the necessity of integrating truncated multi-step targets to effectively manage shifting bias.

\bibliography{TR-CS-UOM-2023}
\bibliographystyle{TR-CS-UOM-2023}
\end{document}